\documentclass[final,5p,times,twocolumn,preprint]{elsarticle}

\usepackage{amssymb}

\usepackage{epsfig} 
\usepackage{CJK}
\usepackage{graphicx}
\usepackage{caption}
\usepackage{amsmath}
\usepackage{float}
\usepackage{epsfig}
\usepackage{url}
\usepackage{verbatim}
\usepackage{amssymb}
\usepackage{fancyhdr}
\usepackage{subfigure}
\usepackage{booktabs}
\usepackage{xcolor}
\usepackage{textcomp}
\usepackage{algorithm}
\usepackage{amsmath}
\usepackage{multirow}
\usepackage{algorithm}
\usepackage{algorithmic}

\journal{Knowledge-Based Systems}

\begin{document}

\begin{frontmatter}



\title{Multi-view Data Classification with a Label-driven Auto-weighted Strategy}

\author[a]{Yuyuan Yu}
\ead{isyuyuyuan@hotmail.com}
\author[a,b]{Guoxu Zhou \corref{*}}
\ead{gx.zhou@gdut.edu.cn}
\fntext[*]{* Corresponding author.}
\author[a]{Haonan Huang}
\ead{mrhaonan@aliyun.com}
\author[a,d]{Shengli Xie}
\ead{shlxie@gdut.edu.cn}
\author[a,c,e]{Qibin Zhao}
\ead{qibin.zhao@riken.jp}

\address[a]{School of Automation, Guangdong University of
	Technology, Guangzhou 510006, China}
\address[b]{Key Laboratory of Intelligent Detection and The Internet of Things in Manufacturing, Ministry of Education, Guangzhou 510006, China}
\address[c]{Joint International
	Research Laboratory of Intelligent Information Processing and System Integration of IoT, Ministry of Education, Guangdong University of Technology,
	Guangzhou 510006, China}
\address[d]{Guangdong-Hong Kong-Macao Joint Laboratory for Smart Discrete Manufacturing, Guangdong University of Technology, Guangzhou 510006, China}
\address[e]{Center for Advanced Intelligence
	Project (AIP), RIKEN, Tokyo, 103-0027, Japan}

\begin{abstract}
Distinguishing the importance of views has proven to be quite helpful for semi-supervised multi-view learning models. However, existing strategies cannot take advantage of semi-supervised information, only distinguishing the importance of views from a data feature perspective, which is often influenced by low-quality views then leading to poor performance. In this paper, by establishing a link between labeled data and the importance of different views, we propose an auto-weighted strategy to evaluate the importance of views from a label perspective to avoid the negative impact of unimportant or low-quality views. Based on this strategy, we propose a transductive semi-supervised auto-weighted multi-view classification model. The initialization of the proposed model can be effectively determined by labeled data, which is practical. The model is decoupled into three small-scale sub-problems that can efficiently be optimized with a local convergence guarantee. The experimental results on classification tasks show that the proposed method achieves optimal or sub-optimal classification accuracy at the lowest computational cost compared to other related methods, and the weight change experiments show that our proposed strategy can distinguish view importance more accurately than other related strategies on multi-view datasets with low-quality views.
	
\end{abstract}


\end{frontmatter}

\section{Introduction}
In scientific fields such as pattern recognition, biomedicine, and data mining, a vast amount of heterogeneous data has evolved as a result of the advancement of data collection technology \cite{zhang2021cmc}. These data, also known as multi-view data, are created using several data gathering sources or feature construction methods. For example, in visual data, each image and video can be built with multiple views from different visual features such as CENT, CMT, GIST, HOG, LBP, and SIFT; in biomedical data, it can measure genes using various techniques such as gene expression, single-nucleotide polymorphism, methylation \cite{jiang2019lightcpg}, and array comparative genomic hybridization; for news text data, it can characterize it in a variety of languages \cite{zhang2018generalized}. Learning a single view may be sufficient for the basic needs of some tasks, such as clustering \cite{liang2020multi}, semi-supervised classification \cite{wang2021seeded}, etc. However, the performance of the final task is frequently improved by properly combining multiple views of the data and learning the consensus representation \cite{chen2021multiview,zhang2021deep,zhang2017latent}. As a result, multi-view learning has received increasing research attention.

Transductive semi-supervised multi-view learning models try to learn the consensus representation using a small quantity of multi-view labeled data and a big amount of multi-view unlabeled data \cite{li2010two}. To exploit labeled data, the models commonly use the following techniques: (1) assuming that the data is uniquely on a low-dimensional manifold, propagating labeled information from labeled data to unlabeled data by label propagation technology \cite{li2017multi}; (2) restricting labeled data with the same label to have the same or similar features in the consensus representation \cite{wang2016adaptive}.

Label propagation technology has become popular in semi-supervised multi-view learning models in recent years. Liang et al. \cite{liang2021semi} proposed to propagate label information over an auto-weighted adaptive neighbor graph and then combing multi-view non-negative matrix decomposition to generate a non-negative consensus representation with highly discriminative. To build a more robust adaptive neighborhood graph, Bo et al. \cite{bo2019latent} proposed to learn latent representations in multi-view data and incorporate label information into the model using label propagation techniques. Bahrami et al. \cite{bahrami2021joint} proposed a multi-view learning method for graph fusion and label propagation that predicts the labels of unlabeled data while fusing graphs. However, there are two major limitations to such these methods that must be overcome \cite{cai2013multi}: on the one hand, as the key to the success of these methods is to construct appropriate similarity graphs and propagate the label information over the graph, different similarity measure functions or hyper-parameters will drastically alter the performance of the final task. Constructing similarity graphs from multi-view data, on the other hand, is a high computational task that is not suitable for large-scale problems.

In semi-supervised multi-view learning, the approach of mapping label information directly to the unified representation is also widely employed. Cai et al. \cite{cai2019semi} proposed a semi-supervised multi-view model based on non-negative matrix decomposition that combines label information through label mapping and integrates co-regularization and sparse restrictions to provide robust consensus representation. Cai et al. \cite{cai2020Semi} introduced a semi-supervised multi-view model based on orthogonal non-negative matrix decomposition, which integrates label mapping, co-regularization, and orthogonal constraints to improve the discriminative power of consensus representation.

Existing semi-supervised multi-view learning methods may effectively learn discriminative consensus representations from multi-view data using label information. However, in some cases, unimportant views can reduce the discriminative power of the model's extracted consensus representations, so treating all views equally is not ideal \cite{tao2017scalable}. To distinguish the importance of different views, there has been some work to investigate adaptively adjusting the weights for the views. Cai et al. \cite{cai2013multi} developed a robust multi-view k-mean clustering method that adaptively adjusts the weight of view based on the fitting error of the view fits and measures the data loss by using the $L_{2,1}$ norm to achieve a robust and stable result. Nie et al. \cite{nie2020auto} proposed a fast multi-view bilateral k-mean clustering method that its features and samples of the data are clustered simultaneously by two indicator matrices, with the weights of different views adaptively adjusted based on the fitting error.

Currently, auto-weighted multi-view learning methods alter the weight of each view adaptively based on the fitting of different views to distinguish the importance of different views. Since this approach establishes a link between view fitting and view importance from the perspective of data features \cite{cai2013multi,nie2020auto}, it can be called a data-driven auto-weighted strategy. This heuristic strategy is adopted by many unsupervised and semi-supervised multi-view learning methods but still has the following limitations: on the one hand, view fitting can be disturbed by many factors, such as model hyper-parameters (e.g., rank or regularized coefficients) and the quality of views; on the other hand, If the views have different data distribution or feature dimensions, the strategy will favor more compact data distribution and feature dimensionality views with smaller orders of magnitude. Both of these limitations can cause the strategy to fail to correctly distinguish the importance of views in some cases, as well as impair the discriminative ability of the representations extracted by the methods. Therefore, how to correctly distinguish the importance of views remains an imminent and urgent problem. In this paper, we consider the importance of views from a labeling perspective to address this problem in the semi-supervised field. A key factor in distinguishing the views from a label perspective is associating the label information of data with individual views. Since the label information does not differ across views but is only associated with the data itself. So trying to measure the importance of views from a labeling perspective is actually a tricky task.

In this paper, we propose a label-driven auto-weighted strategy. Based on this strategy, we propose a transductive semi-supervised multi-view classification method, called label-driven automatic weighted constrained K-means (LACK) method, is also proposed. The highlights and main contributions of this paper are summarized as follows:
\begin{itemize}
\item [(1)] To the best of our knowledge, this is the first auto-weighted strategy that distinguishes the importance of views for the multi-view dataset from the label perspective.
\item [(2)] Compared with the existing auto-weighted strategy, the proposed strategy distinguishes the importance of views more accurately at a lower computational cost, especially when low-quality views exist in a multi-view dataset. Moreover, the proposed strategy is extremely scalable.
\item [(3)] A transductive semi-supervised multi-view classification model based on our auto-weighted strategy is also proposed, which is parameter-free. The initialization of the proposed model can be effectively fixed with label information, which is extremely practical.
\item [(4)] The model can be decoupled into three small-scale sub-problems to achieve convergence in linear time. Complexity analysis and convergence analysis of our method are provided.
\end{itemize}
\begin{figure*}
	\label{fig:A1}
	\includegraphics[width=1\linewidth]{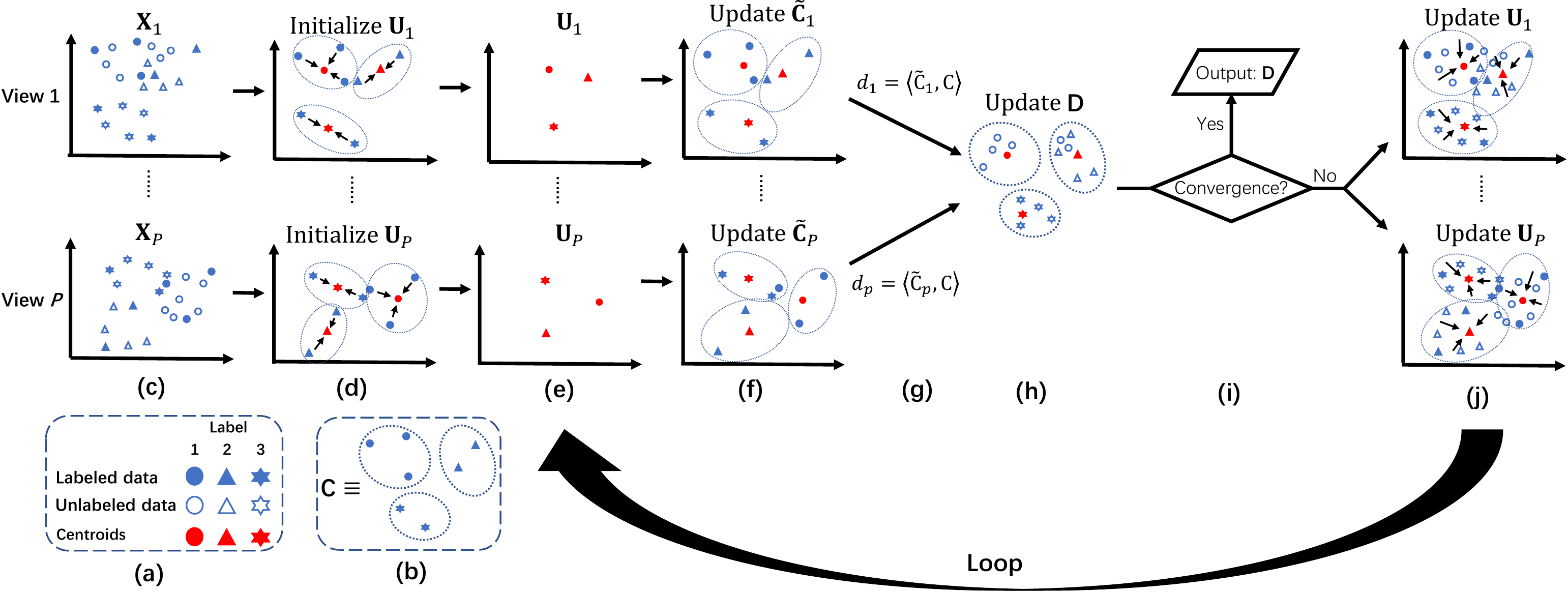}
	\caption{Flowchart of LACK. (a) Graphical representation of labeled data, unknown labeled data and centroids; (b) The true label matrix $\mathbf{C}$; (c) Given the multi-view data $\mathbf{X}_{p}$ with $P$ views, where $p=1, \ldots, P$; (d) Initialize the centriod matrix $\mathbf{U}_{p}$ by the labeled data only to maximize the use of label information; (e) Obtain the centroid matrix $\mathbf{U}_{p}$; (f) Partition the label data by the current centroid $\mathbf{U}_{p}$ to Update the indicator matrix $\widetilde{\mathbf{C}}_{p}$; (g) Compare the difference between $\widetilde{\mathbf{C}}_{p}$ and $\mathbf{C}$ to measure the quality of the $p$th centriod and update the weight $d_{p}$; (h) By adaptively adjust the weights and take full advantage the relationship of multi-view data to learn the indicator matrix $\mathbf{D}$ of the unlabeled data; (i) Determine if the convergence condition is true, output $\mathbf{D}$ if it is true, and continue with step "j" if it is not true; (j) Update the centroid matrix $\mathbf{U}_{p}$ with the partition for all data and then return to step "e".}
\end{figure*}
\section{Preliminaries}
\label{sec:2}
\subsection{Notations}
\label{sec:2.1}
The notations used in this paper are summarized in this subsection. For a multi-view dataset, $P$ denotes the total number of views, $p=1, \ldots P$ denotes the index of view, $\mathbf{X}_{p} \in \mathbb{R}^{m_{p}\times n}$ denotes the $p$th view, where $m_{p}$ denotes the dimension of $p$th-view data features and $n$ denotes the number of data samples. The $i$th row and $j$th column of $\mathbf{X}_{p}$ can be represented by $x^{p}_{i \cdot}$ and $x^{p}_{\cdot j}$, respectively. $\Phi$ is defined as the indicator matrix set which stores the clustering indicator. For the indicator matrix $\mathbf{Q} \in \Phi_{n \times c}$, each row of the matrix $\mathbf{Q}$ has only one element of 1 and all others are 0, then the column corresponding to element of 1 is the clustering indicator of the sample, where $c$ denotes the number of categories of data.

\subsection{Related Work}
\label{sec:2.2}
As one of the most famous clustering method, K-means is widely used for large-scale data clustering due to its effectiveness and simplicity. But K-means can only perform clustering to single view. To solve the multi-view data clustering problem, the Naive Multi-view K-means Clustering model can be naturally developed as follows:
\begin{equation}
	\label{naive_MK}
	\begin{aligned}
		&\min_{\mathbf{U}_{p},\mathbf{Q}}\sum_{p=1}^{P}\left\|\mathbf{X}_{p}-\mathbf{U}_{p} \mathbf{Q} \right\|^{2}_{F}\\
		&\text{s.t.}\ \mathbf{Q} \in \Phi_{c \times n},
	\end{aligned}
\end{equation}
where $\mathbf{U}_{p}\in\mathbb{R}^{m_{p} \times c}$ denotes the centroid matrix of the $p$th-view, $\mathbf{Q}$ denotes the indicator matrix and $\mathbf{Q}$ denotes a sparse matrix.

It is well known that traditional multi-view learning methods aim to learn consensus information from multi-view data. However, this is not optimal, because they treat important views and less important views equally and do not take into account the heterogeneity between views. Cai et al. \cite{cai2013multi} proposed a Robust Large-scale Multi-view K-means Clustering (RMVKM) method which adaptively learns the weight of view to differentiate the importance of views by a data-driven auto-weighted strategy. The RMVKM problem can be defined as follow:
\begin{equation}
	\begin{aligned}
		&\min_{\mathbf{U}_{p},\mathbf{Q}, d_{p}}\sum_{p=1}^{P}({d_{p}})^{\gamma}\left\|\mathbf{X}_{p}-\mathbf{U}_{p} \mathbf{Q} \right\|_{2,1}\\
		&\text{s.t.} \mathbf{Q} \in \Phi_{c \times n}, \sum_{p=1}^{P} d_{p}=1, d_{p} \geq 0,
	\end{aligned}
\end{equation}
where $d_{p}$ denotes the weight for the $p$th-view and $\gamma$ denotes the trade-off parameter to control the weights distribution. $d_{p}$ can be obtained by
\begin{equation}
	\label{weight_RMVKM}
	d_{p}=\frac{(\gamma e_{p})^{\frac{1}{1-\gamma}}}{\sum_{p=1}^{P}(\gamma e_{p})^{\frac{1}{1-\gamma}}},
\end{equation}
where $e_{p}=\left\|\mathbf{X}_{p}-\mathbf{U}_{p} \mathbf{Q}^{\top} \right\|_{2,1}$ denotes the fitting error of the $p$th view. $\gamma$ is used to manually adjust the weight distribution to obtain better weights. However, manually searching for the best gamma is costly and impractical. The recent research \cite{nie2020auto,nie2017learning,huang2020auto} use a parameter-free auto-weighted strategy to obtain $d_{p}$ and proves its effectiveness. Using the RMVKM method as an example, $d_{p}$ can be obtained through the new strategy in the following:

\begin{equation}
	\label{weight_data}
	d_{p}=\frac{1}{2e_{p}}.
\end{equation}

Since $d_{p}$ is related to the $p$th view data fitting, this strategy is called data-driven auto-weighted strategy in this paper.

\section{The proposed method}
\label{sec:3}
In this section, we first construct a base model called Semi-supervised Multi-view Classification via Constrained K-means (MLCK). Second, we proposed a label-driven auto-weighted strategy to distinguish the importance of views. Thirdly, by combining the strategy with the MLCK model, we proposed a semi-supervised multi-view classification model called Label-driven Auto-weighted Constrained K-means (LACK). Then, we decoupled the LACK model into three small-size problems for efficient alternating optimization. Finally, we performed the computational complexity analysis and convergence analysis of LACK.
\subsection{LACK model}
Let $\{\mathbf{X}_{p}\}_{p=1}^{P} \in \mathbb{R}^{m_{p} \times n}$ denotes the multi-view dataset with $P$ views, where $m_p$ denotes the feature dimension of $p$th-view and $n$ denotes the number of samples. Considering $\{\mathbf{X}_{p}\}_{p=1}^{P}$ consisting of $n$ data points $\{x_{i}\}_{i=1}^{n}$, where the first $l$ data points have the label information and the remaining $n-l$ data points are unlabeled, then the label constraint matrix $\mathbf{Q} \in \Phi_{c \times n}$ can be defined as follows:
\begin{equation}
	\begin{aligned}
		\label{initial_Q}
		\mathbf{Q}=\left[ \mathbf{C}, \mathbf{D} \right],
	\end{aligned}
\end{equation}
where $\mathbf{C} \in \Phi_{c \times l}$ denotes the true label matrix. If $x_{i}$ is labeled with the $j$th class, then $c_{ji}=1$ and the other elements of $c_{\cdot i}$ are $0$; and $\mathbf{D} \in \Phi_{c \times (n-l)}$ denotes the indicator matrix. For example, consider $n$ data points with 3 class, among which $x_{1}$, $x_{2}$ are labeled with class $1$, $x_{3}$, $x_{4}$ are labeled with class $2$, $x_{5}$, $x_{6}$ are labeled with class $3$, and the other $n-6$ data points are unlabeled. Based on above example, $\mathbf{Q}$ can be represented as follows:
\begin{equation}
	\label{obtain_Q}
	\begin{aligned}
		\mathbf{Q}=&\left[ \mathbf{C}, \mathbf{D} \right]\\=&\left[\begin{array}{cccccccccc} 
			1 & 1 & 0 & 0 & 0 & 0 & 0 &\cdots & 0\\ 
			0 & 0 & 1 & 1 & 0 & 0 & 0 &\cdots & 0\\ 
			0 & 0 & 0 & 0 & 1 & 1 & 0 &\cdots & 0
		\end{array}\right].
	\end{aligned}
\end{equation}

To handle the transductive semi-supervised multi-view learning problem, we constructed the MLCK model by incorporating the label information in Eq.\eqref{naive_MK} as follows
\begin{equation}
	\label{problem_MLCK}
	\begin{aligned}
		&\min_{\mathbf{U}_{p},\mathbf{D}}\sum_{p=1}^{P}\left\|\mathbf{X}_{p}-\mathbf{U}_{p} \mathbf{Q} \right\|^{2}_{F}\\
		&\text{s.t.} \mathbf{U}_{p}\in\mathbb{R}^{m_{p} \times c}, \mathbf{Q}=\left[ \mathbf{C}, \mathbf{D} \right] \in \Phi_{c \times n}.
	\end{aligned}
\end{equation}
%
%
%

MLCK can be considered as a multi-view extended version of Constrained K-means (CK) \cite{basu2002semi}. The only difference is that MLCK learns the indicator matrix from multiple views of the data. However, MLCK ignores the importance of different views, which is not optimal. Therefore, we develop the follows by combining the auto-weighted strategy:
\begin{equation}
	\begin{aligned}
		\label{LACK}
		&\min_{\mathbf{U}_{p},\mathbf{D}}\sum_{p=1}^{P}d_{p}\left\|\mathbf{X}_{p}-\mathbf{U}_{p} \mathbf{Q}\right\|^{2}_{F}\\
		&\text{s.t.} \mathbf{U}_{p}\in\mathbb{R}^{m_{p} \times c}, \mathbf{Q}=\left[ \mathbf{C}, \mathbf{D} \right] \in \Phi_{c \times n},
	\end{aligned}
\end{equation}
where $d_{p}$ denotes the weight of the $p$th-view. 

According to the data-driven auto-weighted strategy \cite{nie2020auto}, $d_{p}$ can be obtained by 

\begin{equation}
	\begin{aligned}
		\label{DACK}
		d_{p}=\frac{1}{2\left\|\mathbf{X}_{p}-\mathbf{U}_{p} \mathbf{Q}\right\|_{F}},
	\end{aligned}
\end{equation}
where the value of $d_{p}$ is determined by the data fitting of the $p$th view, and the better view fitting, the larger $d_{p}$ is.

However, the data-driven auto-weighted strategy Eq.\eqref{DACK} still has two main drawbacks: on the one hand, the view fitting is often susceptible to the interference of various factors, such as model hyper-parameters (e.g., rank or regularized coefficients) and the quality of views; on the other hand, the strategy would favor views with compact data distribution and smaller features dimensions. These shortcomings cause data-driven strategies to fail to accurately distinguish the importance of views in some cases.

To solve this problem, we propose a label-driven auto-weighted strategy using label information as a prior information as follows:
\begin{equation}
	\label{weight_new}
	d_{p}=\left \langle \widetilde{\mathbf{C}}_{p}, \mathbf{C} \right \rangle,
\end{equation}
where $\left \langle \cdot,\cdot \right \rangle$ denotes the inner product and $\widetilde{\mathbf{C}}_{p} \in \Phi_{c \times l}$ denotes the label prediction matrix for the labeled data of the $p$th-view. If $\widetilde{\mathbf{C}}_{p}$ is very similar to the true label matrix $\mathbf{C}$, i.e., $\mathbf{U}_{p}$ clusters most of the labeled data into the correct class, then the $p$th view is more important and can obtain a larger value of $d_{p}$ by Eq.\eqref{weight_new}.

The $i$th row of $\widetilde{\mathbf{C}}_{p}$ can be updated by solving the following:
\begin{equation}
	\label{weight_new_problem}
	\min_{\widetilde{\mathbf{c}}_{i \cdot},  i=1,\ldots,l}\left\|x^{p}_{\cdot i}-\mathbf{U}_{p}\widetilde{\mathbf{c}}_{\cdot i}\right\|^{2}_{2},
\end{equation}
where $x^{p}_{\cdot i}\in \mathbb{R}^{m_{p} \times 1}$ denotes the $i$th label data of the $p$th-view and $\widetilde{\mathbf{c}}_{\cdot i} \in \Phi_{c \times 1}$ denotes the indicator vector of $x^{p}_{\cdot i}$ assigned by the centroid matrix $\mathbf{U}_{p}$.

Recall that the purpose of the auto-weighted strategy aims to distinguish the importance of different views to learn a better consensus representation \cite{nie2017auto,chen2019auto}. It is clear that the bridge connecting the view to the indicator matrix (a.k.a. the consensus representation of data) is the centroid matrix in our model. Therefore, to obtain a reliable indicator matrix, our proposed strategy adjust the weight of view by assessing the quality of the centroid matrix. If the learned centroid matrix of the view can cluster the labeled data more correctly, then it indicates that this view data is important and needs to be given huge weight. We achieve the above idea by solving the problem \eqref{weight_new_problem} and calling it a label-driven auto-weighted strategy. Compared with the traditional auto-weighted strategy, our proposed strategy distinguish the importance of views more accurately.

\subsection{Optimization Algorithm}
In this subsection, we proposed a optimization algorithm to updates $\mathbf{U}_{p}$, $d_{p}$ and $\mathbf{D}$ for the problem ($\ref{LACK}$), respectively. When updating one of these variables, other variables need to be fixed. It should be noted that the optimization algorithm emphasizes fastness when updating variables. For example, A fast algorithm \cite{nie2020auto} on sparse matrix multiplication is used to efficiently solve $\mathbf{U}_{p}$, a fast algorithm \cite{cai2013multi} that avoids huge matrix multiplications is used to efficiently solve $\mathbf{D}$ and $d_{p}$. Our proposed optimization algorithm has a great positive effect on the fastness of the algorithm compared to other complicated solution processes. Below, we will show the specific steps of this optimization algorithm.

\subsubsection{Initialization} 
We first initialize the variables $\mathbf{D}$ as a zeros matrix and initialize the true label matrix $\mathbf{C}$ based on the label information, and then the label constraint matrix $\mathbf{Q}$ can be obtained uniquely similar to Eq.\eqref{obtain_Q}. The initial centroid matrix $\mathbf{U}_{p}$ can be constructed by $\mathbf{U}_{p}=\mathbf{X}_{p} \mathbf{Q}^{\top}(\mathbf{Q}\mathbf{Q}^{\top})^{-1}$. This initialization method makes MLCK avoid the unstable classification performance caused by random initialization of the centroid matrix, which greatly improves the practicality.

\subsubsection{Update $\mathbf{U}_{p}$} Solving $\mathbf{U}_{p}$, when $\mathbf{Q}$ and $d_{p}$ are fixed.

The problem \eqref{LACK} can be rewritten as:
\begin{equation}
	\label{objective_U}
	\begin{aligned}
		J=&\min_{\mathbf{U}_{p}}d_{p} \operatorname{Tr}\{(\mathbf{X}_{p}-\mathbf{U}_{p}\mathbf{Q})(\mathbf{X}_{p}-\mathbf{U}_{p}\mathbf{Q})^{\top}\}\\
		=&\min_{\mathbf{U}_{p}} d_{p}\{\operatorname{Tr}(\mathbf{X}_{p}\mathbf{X}_{p}^{\top})-2\operatorname{Tr}(\mathbf{U}_{p}\mathbf{Q}\mathbf{X}_{p}^{\top})\\
		&+\operatorname{Tr}(\mathbf{U}_{p}\mathbf{Q}\mathbf{Q}^{\top}\mathbf{U}_{p}^{\top})\}.
	\end{aligned}
\end{equation}

By deriving the derivative of $J$ with respect to $\mathbf{U}_{p}$ and setting to 0, we have:
\begin{equation}
	\begin{aligned}
		\frac{\partial J}{\partial \mathbf{U}_{p}}&=-2\mathbf{X}_{p}\mathbf{Q}^{\top}+2\mathbf{U}_{p}\mathbf{Q}\mathbf{Q}^{\top}\\
		&=2(\mathbf{U}_{p}\mathbf{Q}-\mathbf{X}_{p})\mathbf{Q}^{\top}=0.
	\end{aligned}
\end{equation}

The solution of problem \eqref{objective_U} can be obtained by

\begin{equation}
	\label{update_U}
	\mathbf{U}_{p}=\mathbf{Q}^{\top}(\mathbf{Q}\mathbf{Q}^{\top})^{-1}.
\end{equation}

As each row of $\mathbf{Q} \in \Phi_{c \times n}$ has only one element of 1 and all other are 0, and $\mathbf{Q}\mathbf{Q}^{\top} \in \mathbb{R}^{c \times c}$ is a diagonal matrix, $\mathbf{U}_{p}$ can be easily obtained.

\subsubsection{Update $d_{p}$} Solving $d_{p}$, when $\mathbf{U}_{p}$ and $\mathbf{D}$ are fixed. 

For the problem \eqref{weight_new_problem}, the solution $\widetilde{\mathbf{c}}_{i \cdot}$ can be obtained by
\begin{equation}
	\label{update_c}
	\widetilde{\mathbf{c}}_{i j}=
	\begin{cases}
		1,\quad j=\underset{k}{\arg\min}\sum_{p=1}^{P}d_{p}\left\|x^{p}_{\cdot i}-u^{p}_{\cdot k}\right\|^{2}_{2}\
		\\
		0,\quad otherwise.
	\end{cases}
\end{equation}

By solving $l$ minimization problem \eqref{weight_new_problem} using \eqref{update_c}, $\widetilde{\mathbf{C}}_{p}$ can be obtained and then the closed-form solution of $d_{p}$ can be obtained by Eq. \eqref{weight_new}.

\subsubsection{Update $\mathbf{D}$} Solving $\mathbf{D}$, when $\mathbf{U}_{p}$ and $d_{p}$ are fixed.

The problem \eqref{LACK} can be rewritten as follow:

\begin{equation}
	\begin{aligned}
		&\min_{\mathbf{D}}\sum_{p=1}^{P}d_{p}\left\|\mathbf{X}_{p}-\mathbf{U}_{p} \mathbf{Q} \right\|^{2}_{F}\\
		=&\min_{q_{\cdot i},n-l \leq i \leq n}\sum_{p=1}^{P}\sum_{i=l+1}^{n}d_{p}\left\|x^{p}_{\cdot i}-\mathbf{U}_{p}q_{\cdot i}\right\|^{2}_{2}.
	\end{aligned}
\end{equation}
where $x^{p}_{\cdot i} \in \mathbb{R}^{m_{p} \times 1}$ denotes the $i$th sample of $p$th-view and $q_{\cdot i} \in \Phi_{c \times 1}$ denotes the indicator of $i$th sample. For different variable $q_{\cdot i}$ in the $n-l \leq i \leq n$ cases, the solution process is independent of each other. Therefore, we can obtain $q_{\cdot i}$ by solving the following problem:

\begin{equation}
	\min_{{q_{\cdot i},n-l \leq i \leq n}}\sum_{p=1}^{P}d_{p}\left\|x^{p}_{\cdot i}-\mathbf{U}_{p}q_{\cdot i}\right\|^{2}_{2}.
\end{equation}

For the indicator $q_{\cdot i}$ of the $i$th sample, only one of its elements is equal to 1 and the rest are 0. Hence, $q_{j i}$ can be obtained by

\begin{equation}
	\label{update_Q}
	q_{ji}=
	\begin{cases}
		1,\quad j=\underset{k}{\arg\min}\sum_{p=1}^{P}d_{p}\left\|x^{p}_{\cdot i}-u^{p}_{\cdot k}\right\|^{2}_{2}\
		\\
		0,\quad otherwise,
	\end{cases}
\end{equation}
where $u^{p}_{\cdot k} \in \mathbb{R}^{m_{p} \times 1}$ denotes the $k$th column of $\mathbf{U}_{p}$.

Based on the above analysis, we propose the LACK method, whose procedure is shown in Alg. \ref{LACK_algorithm}, where the initialization of $\mathbf{U}_{p}$ was omitted because $\mathbf{U}_{p}$ is updated in the same way as it is initialized method and is the first updated variable in each view.

\begin{algorithm}[tbp] \small
	\caption{LACK}
	\label{LACK_algorithm}
	\begin{algorithmic}[1]
		\REQUIRE Multi-view data $\mathcal{X}_{p} \in \mathbb{R}^{m_{p} \times n}$ with $p$th-view, the true label indicator matrix $\mathbf{C} \in \Phi_{l \times c}$ the number of labels $c$.\\ 
		\ENSURE The indicator matrix $\mathbf{D} \in \Phi_{(n-l) \times c}$.\\
		\STATE{Initialize $\mathbf{Q}$ by Eq.\eqref{initial_Q}} and $d_{p}=1/p$.
		\REPEAT 
		\FOR{$p$ = 1 to $P$}
		\STATE Update $\mathbf{U}_{p}$ by Eq.\eqref{update_U}.
		\STATE Update $\widetilde{\mathbf{C}}_{p}$ by Eq.\eqref{update_c}.
		\STATE Update $d_{p}$ by Eq.\eqref{weight_new}.
		\ENDFOR\\
		\STATE Update $\mathbf{D}$ by Eq.\eqref{update_Q}.\\
		\UNTIL{convergence}.
	\end{algorithmic}
\end{algorithm}
\subsection{Computational Complexity Analysis}
\label{Computation}
In this subsection, we discuss the computation complexity of our proposed LACK method. For simplicity, we first set the following conditions: $n$ denotes the number of samples, $m$ denotes the feature dimension for each view, $P$ denotes the number of views, $t$ denotes the number of iterations, $c$ denotes the number of labels and $l$ denotes the number of label examples. According to Alg. \ref{LACK_algorithm}, it can see that $\mathbf{U}_{p}$, $\widetilde{\mathbf{C}}_{p}$, $d_{p}$ and $\mathbf{Q}$ need to be updated. 
The computational complexity for updating $\mathbf{U}_{p}$, $\widetilde{\mathbf{C}}_{p}$, $d_{p}$ and $\mathbf{Q}$ are $O(npt+cpt+ncpt+mnpt)$, $O(mlcpt)$, $O(lpt)$ and $O((m+1)(n-l)cpt)$, respectively. Therefore, the overall computational complexity of LACK is $O(mncpt)$, which is friendly to large-scale multi-view data.

The computational complexity of our proposed label-driven auto-weighted strategy is $O(mlcpt+lpt)$, which is much less than that of $O(mnpt)$ for the data-driven auto-weighted strategy by Eq.\eqref{DACK}. It can be known that our proposed strategy is more effective than the traditional strategy when dealing with the large-scale problem with a small number of labels, i.e., $n \gg cl$.

\subsection{Convergence Analysis}
\label{convergence}
The convergence of the LACK method can be proved. According to Alg.\ref{LACK_algorithm}, the problem \eqref{LACK} can be decomposed into three sub-problems and each sub-problem is a convex problem with respect to one variable. Hence, by solving the sub-problems alternately, LACK will guarantee that we can find the optimal solution for each sub-problem, and finally, LACK will converge to a local solution \cite{cai2013multi}.

\section{Experimental Analysis}
\label{sec:4}
In this section, we evaluate the performance of our proposed LACK method on the benchmark datasets. All experiments are performed on a Windows 10 machine with an i7 CPU at 2.80 GHz and 16 GB memory.
\subsection{Comparative Methods}
To show the performance of LACK, the state-of-the-art multi-view learning methods for clustering and semi-supervised classification were used as the comparison methods:
\begin{itemize}
	\item NMKC \cite{xu2017re}: Naive Multi-view K-means Clustering (NMKC) learns a unified consensus clustering indicator matrix by integrating heterogeneous information from different perspectives, which can be seen as a multi-view version of K-means.
	\item RMKMC \cite{cai2013multi}: RMKMC is a multi-view K-means method that enhances model robustness through sparsely structured induced $L_{2,1}$-norm and explores different views importance using a data-driven auto-weighted strategy.
	\item LMVSC \cite{kang2020large}: LMVSC integrates complementary information from different views through a unified anchor graph, and thus implements spectral clustering with linear complexity on this anchor graph.
	\item CK \cite{basu2002semi}: Constrained K-means (CK) is a semi-supervised single-view classification method that efficiently uses label information to initialize the centroid matrix and facilitate the generation of indicator matrix for unlabeled data. To handle the multi-view dataset, the features of all views are concatenated as input to CK.
	\item MVCNMF \cite{cai2019semi}: MVCNMF is a semi-supervised multi-view model based on NMF. It integrates complementary information from different views through co-regularization and sparse constraints to obtain robust low-dimensional features.
	\item MVOCNMF \cite{cai2020Semi}: MVOCNMF is a semi-supervised multi-view model based on orthogonal NMF. It enhances the discriminative power of consensus representation by integrating complementary information of views.
	\item LMSSC \cite{bo2019latent}: LMSSC dynamically constructs graphs and performs label propagation. It can find the potential consensus representations by explore potential complementary information from different views.
	\item AMGL \cite{nie2016parameter}: AMGL is a parameter-free auto-weighted multiple graph learning method to learn a fusion graph. It distinguishes the importance of views by the data-driven auto-weighted strategy. We chose the semi-supervised version of AMGL as our comparison method.
	\item MLAN \cite{nie2017auto}: MLAN performs both semi-supervised multi-view learning and local structure learning and distinguishes the importance of views by a data-driven auto-weighted strategy to obtain the optimal graph.
	\item MLCK:MLCK is a semi-supervised multi-view classification method, equivalent to a multi-view version of the CK model, which learns an independent cluster centroid for each view data and an indicator matrix, where all views are treated equally.
	\item DACK: DACK is a semi-supervised multi-view classification method, which is equivalent to MLCK combined with a data-driven auto-weighted strategy Eq.\eqref{weight_data} to learn a more reliable indicator matrix.
\end{itemize}
\subsection{Datasets and Experiment Setting}
\begin{table}[!t] \scriptsize
	\renewcommand{\arraystretch}{1.2}
	\caption{Descriptions of the datasets.}
	\centering
	\label{table:dataset}
	\setlength{\tabcolsep}{0.6mm}
	\begin{tabular}{c|ccc}
		\hline
		\hline
		View & Caltech-101-7 & Caltech-101-20 & MSRC-v1 \\ \hline
		1 & GABOR (48) & GABOR (48) & CM (24)\\
		2 & WM (40) & WM (40) & HOG (576)\\
		3 & CEMTRIST (254) & CEMTRIST (254) & GIST (512) \\
		4 & HOG (1984) & HOG (1984) & LBP (256)\\
		5 & GIST (512)  & GIST (512)& CENTRIST (254)\\
		6 & LBP (928)  & LBP (928)& $-$\\
		Size & 1474 & 2386 & 210\\
		Classes & 7 & 20 & 7\\ 
		\hline
		\hline
	\end{tabular}
\end{table}
The details of three benchmark datasets that were used to evaluate the performance of LACK are shown in Table \ref{table:dataset}.
\begin{itemize}
	\item Caltech-101-7(20) \cite{dueck2007non}: Caltech-101 is an object recognition dataset containing 8,677 images in 10 categories. Following the work \cite{nie2020auto}, we selected the commonly used benchmark subsets Caltech-101-7 and Caltech-101-20. Each image in the dataset is extracted with six visual feature vectors to compose six views of the data, namely, the Gabor feature of dimension 48, Wavelet moments (WM) of dimension 40, the CENTRIST feature of dimension 254, the HOG feature of dimension 1984, the GIST feature of dimension 512 and the LBP feature of dimension 928.
	\item MSRC-v1 \cite{winn2005locus}: This is a scene recognition dataset containing 240 images in 8 categories. Based on the recent work \cite{wang2021learning}, seven categories were adopted, which are airplane, bicycle, building, car, cow, face, and tree. Five visual feature vectors are extracted from each image, which are the CM feature of dimension 24, the HOG feature of dimension 576, the GIST feature of dimension 512, the LBP feature of dimension 256, and the CENTRIST feature of dimension 254.
\end{itemize}

In the experiments, four evaluation metrics were used to evaluate the effectiveness of the methods, including Accuracy (ACC), F-score, Precision, and Recall. For additional details, please refer to the previous work \cite{schutze2008introduction}. It is worth noting that a higher value of the evaluation metric indicates better performance. We perform experiments with training and test samples as a whole to simulate the transductive semi-supervised multi-view learning problem. We denote $\tau$ as the label ratio, $\tau=\{0.01, 0.1\}$ for semi-supervised classification experiments, that is, we take $\operatorname{ceil}(\tau n_{c})$ as the training samples of the $c$-th classes, where $n_{c}$ denotes the total number of samples of the $c$-th category and $\operatorname{ceil} (\cdot)$ denotes the operation of rounding up to an integer in MATLAB. We also compare several multi-view clustering methods for handling large-scale data to highlight the efficiency of our method.

To ensure the fairness of the experiment, we set the parameters of the comparison methods according to the descriptions in the corresponding papers. The iteration termination condition of LACK is set to the difference of variable $\mathbf{Q}$ in two adjacent iterations is zero. In all experiments, each method is run 10 times and reported the average results. The best results in each experiment marked by bold.

\subsection{Semi-supervised Classification Experiments}
In this subsection, we design a semi-supervised classification experiment to compare the classification performance and efficiency of LACK with that of the state-of-the-art transductive semi-supervised multi-view classification methods under the label information ratios $\tau=\{0.01,0.1\}$. In addition, we also compare three unsupervised multi-view algorithms for large-scale data to highlight the efficiency of our algorithm.

The experimental results reported in Tables \ref{table:C1}, \ref{table:C2}, \ref{table:C4} and \ref{table:C5}. It can be obtained the following point: LACK achieved leading or near-optimal performance results in most cases due to its ability to learn multi-view unified indicators in a semi-supervised K-means multi-view framework and adaptively distinguish the importance of views by a label perspective. Moreover, LACK has a substantial efficiency advantage compared to the other multi-view methods due to its low computational complexity, which is only linear in $n$, so it can easily handle data of different scales. In addition, It can be seen that the standard deviation of LACK is zero for each experiment. The reason is that LACK is parameter-free and can initialize the variables fixedly by data and label information. This property also greatly enhances the usefulness of LACK, which saves the time of manually adjusting the optimal parameters and also makes the classification results more interpretable.

\begin{table}[!t] \scriptsize
	\renewcommand{\arraystretch}{1.2}
	\caption{Semi-supervised classification performance of different methods on Caltech-101-7 dataset.}
	\centering
	\label{table:C1}
	\setlength{\tabcolsep}{1mm}
	\begin{tabular}{c|c|cccc}
		\hline
		\hline
		$\tau$ & Method & ACC & F-score & Precision & Recall\\ \hline
		\multirow{3}{*}{$0$}& NMKC & 53.15 $\pm$ 6.64 & 56.56 $\pm$ 5.20 & 89.45 $\pm$ 2.85 & 41.47 $\pm$ 5.05\\ 
		& RMKMC & 53.94 $\pm$ 7.85 & 56.97 $\pm$ 6.37 & 86.10 $\pm$ 7.58 & 42.64 $\pm$ 5.51 \\
		& LMVSC & 60.01 $\pm$ 7.07 & 55.63 $\pm$ 6.20 & 65.54 $\pm$ 4.31 & 48.78 $\pm$ 8.36 \\  \hline
		\multirow{6}{*}{$0.01$}	& CK & 53.53 $\pm$ 0.00 & 60.23 $\pm$ 0.00 & 88.78 $\pm$ 0.00 & 45.58 $\pm$ 0.00 \\
		& MVCNMF & 51.06 $\pm$ 6.57 & 50.29 $\pm$ 5.64 & 74.24 $\pm$ 5.64 & 38.11 $\pm$ 5.18 \\
		& MVOCNMF & 44.33 $\pm$ 8.66 & 45.55 $\pm$ 6.27 & 72.85 $\pm$ 6.55 & 33.21 $\pm$ 5.48 \\
		& AMGL & 84.55 $\pm$ 0.00 & 80.49 $\pm$ 0.00 & 68.35 $\pm$ 0.00 & 97.88 $\pm$ 0.00\\
		& MLAN & \textbf{86.27} $\pm$ \textbf{1.71} & \textbf{82.95} $\pm$ \textbf{2.04} & 71.12 $\pm$ 3.05 & \textbf{99.61} $\pm$ \textbf{0.11}\\
		& LACK & 82.01 $\pm$ 0.00 & 81.31 $\pm$ 0.00 & \textbf{94.81} $\pm$ \textbf{0.00} & 71.17 $\pm$ 0.00 \\
		\hline
		\multirow{6}{*}{$0.1$}& CK & 55.09 $\pm$ 0.00 & 60.48 $\pm$ 0.00 & 86.85 $\pm$ 0.00 & 46.39 $\pm$ 0.00 \\
		& MVCNMF & 53.96 $\pm$ 4.80 & 52.61 $\pm$ 3.02 & 77.65 $\pm$ 4.17 & 39.89 $\pm$ 3.15\\
		& MVOCNMF & 47.29 $\pm$ 5.71 & 48.30 $\pm$ 3.77 & 75.55 $\pm$ 4.81 & 35.59 $\pm$ 3.64\\
		& AMGL & 89.80 $\pm$ 0.00 & 86.74 $\pm$ 0.00 & 77.07 $\pm$ 0.00 & 99.19 $\pm$ 0.00\\
		& MLAN & \textbf{94.50} $\pm$ \textbf{0.04} & \textbf{93.46} $\pm$ \textbf{0.05} & 88.17 $\pm$ 0.09 & \textbf{99.42} $\pm$ \textbf{0.01} \\
		& LACK & 82.01 $\pm$ 0.00 & 82.31 $\pm$ 0.00 & \textbf{97.91} $\pm$ \textbf{0.00} & 71.00 $\pm$ 0.00 \\ 
		\hline
		\hline
	\end{tabular}
\end{table}
\begin{table}[!t] \scriptsize
	\renewcommand{\arraystretch}{1.2}
	\caption{Semi-supervised classification performance of different methods on Caltech-101-20 dataset.}
	\centering
	\label{table:C2}
	\setlength{\tabcolsep}{1mm}
	\begin{tabular}{c|c|cccc}
		\hline
		\hline
		$\tau$ & Method & ACC & F-score & Precision & Recall\\ \hline
		\multirow{3}{*}{$0$}& NMKC & 51.80 $\pm$ 7.07 & 53.05 $\pm$ 10.53 & 81.91 $\pm$ 4.69 & 39.76 $\pm$ 10.90\\ 
		& RMKMC & 50.13 $\pm$ 3.66 & 49.71 $\pm$ 6.15 & 79.82 $\pm$ 3.15 & 36.28 $\pm$ 6.41 \\ 
		& LMVSC & 40.60 $\pm$ 3.75 & 30.86 $\pm$ 3.29 & 36.46 $\pm$ 5.87 & 26.99 $\pm$ 2.75 \\  \hline
		\multirow{6}{*}{$0.01$}& CK & 42.92 $\pm$ 0.00 & 38.55 $\pm$ 0.00 & 71.34 $\pm$ 0.00 & 26.41 $\pm$ 0.00 \\
		& MVCNMF & 45.52 $\pm$ 3.65 & 41.09 $\pm$ 1.99 & 71.68 $\pm$ 3.56 & 28.82 $\pm$ 1.59\\
		& MVOCNMF & 38.76 $\pm$ 2.80 & 34.97 $\pm$ 2.27 & 58.64 $\pm$ 5.45 & 24.94 $\pm$ 1.54\\
		& AMGL & 72.12 $\pm$ 0.00 & 59.74 $\pm$ 0.00 & 43.23 $\pm$ 0.00 & 96.65 $\pm$ 0.00\\
		& MLAN & 74.90 $\pm$ 0.77 & 68.69 $\pm$ 1.82 & 52.88 $\pm$ 2.18 & \textbf{98.09} $\pm$ \textbf{0.10}\\
		& LACK & \textbf{76.24} $\pm$ \textbf{0.00} & \textbf{77.38} $\pm$ \textbf{0.00} & \textbf{90.37} $\pm$ \textbf{0.00} & 67.66 $\pm$ 0.00 \\
		\hline
		\multirow{6}{*}{$0.1$}& CK & 52.68 $\pm$ 0.00 & 56.23 $\pm$ 0.00 & 77.85 $\pm$ 0.00 & 44.00 $\pm$ 0.00 \\
		& MVCNMF & 46.86 $\pm$ 5.15 & 43.25 $\pm$ 5.56 & 71.34 $\pm$ 5.22 & 31.12 $\pm$ 4.97\\
		& MVOCNMF & 41.01 $\pm$ 3.64 & 36.77 $\pm$ 2.64 & 64.92 $\pm$ 5.13 & 25.69 $\pm$ 2.07\\
		& AMGL & 78.32 $\pm$ 0.00 & 69.28 $\pm$ 0.00 & 53.94 $\pm$ 0.00 & 96.80 $\pm$ 0.00\\
		& MLAN & \textbf{84.68} $\pm$ \textbf{0.25} & \textbf{81.18} $\pm$ \textbf{0.43} & 69.44 $\pm$ 0.62 & \textbf{97.71} $\pm$ \textbf{0.04}\\
		& LACK & 76.82 $\pm$ 0.00 & 77.59 $\pm$ 0.00 & \textbf{90.69} $\pm$ \textbf{0.00} & 67.79 $\pm$ 0.00 \\ 
		\hline
		\hline
	\end{tabular}
\end{table}
\begin{table}[!t] \scriptsize
	\renewcommand{\arraystretch}{1.2}
	\caption{Semi-supervised classification performance of different methods on MSRC-v1 dataset.}
	\centering
	\label{table:C4}
	\setlength{\tabcolsep}{1mm}
	\begin{tabular}{c|c|cccc}
		\hline
		\hline
		$\tau$ & Method & ACC & F-score & Precision & Recall\\ \hline
		\multirow{3}{*}{$0$}& NMKC & 75.33 $\pm$ 8.86 & 67.41 $\pm$ 6.87 & 63.69 $\pm$ 8.86 & 71.83 $\pm$ 4.69\\ 
		& RMKMC & 73.52 $\pm$ 9.59 & 66.90 $\pm$ 6.94 & 62.89 $\pm$ 9.28 & 71.86 $\pm$ 4.09\\ 
		& LMVSC & 84.62 $\pm$ 1.57 & 73.93 $\pm$ 2.01 & 72.53 $\pm$ 2.24 & 75.38 $\pm$ 1.84 \\  \hline
		\multirow{6}{*}{$0.01$}& CK & 86.67 $\pm$ 0.00 & \textbf{75.64} $\pm$ \textbf{0.00} & 74.53 $\pm$ 0.00 & \textbf{76.78} $\pm$ \textbf{0.00} \\
		& MVCNMF & 78.82 $\pm$ 3.66 & 66.74 $\pm$ 3.92 & 65.70 $\pm$ 4.48 & 67.83 $\pm$ 3.36\\
		& MVOCNMF & 78.23 $\pm$ 6.32 & 67.01 $\pm$ 7.16 & 65.64 $\pm$ 7.29 & 68.47 $\pm$ 7.11\\
		& AMGL & 74.88 $\pm$ 0.00 & 59.37 $\pm$ 0.00 & 56.86 $\pm$ 0.00 & 62.10 $\pm$ 0.00\\
		& MLAN & 72.51 $\pm$ 3.31& 57.23 $\pm$ 2.74& 52.42 $\pm$ 4.10& 63.15 $\pm$ 0.85\\
		& LACK & \textbf{86.70} $\pm$ \textbf{0.00} & \textbf{75.64} $\pm$ \textbf{0.00} & \textbf{74.81} $\pm$ \textbf{0.00} & 76.50 $\pm$ 0.00 \\
		\hline
		\multirow{6}{*}{$0.1$}& CK & 66.19 $\pm$ 0.00 & 54.40 $\pm$ 0.00 & 53.44 $\pm$ 0.00 & 55.40 $\pm$ 0.00 \\
		& MVCNMF & 77.04 $\pm$ 8.69 & 66.44 $\pm$ 7.90 & 64.62 $\pm$ 9.18 & 68.46 $\pm$ 6.45\\
		& MVOCNMF & 75.93 $\pm$ 4.45 & 66.71 $\pm$ 3.28 & 64.93 $\pm$ 3.65 & 68.61 $\pm$ 3.07\\
		& AMGL & 78.31 $\pm$ 0.00 & 63.16 $\pm$ 0.00 & 60.31 $\pm$ 0.00 & 66.30 $\pm$ 0.00\\
		& MLAN & 83.76 $\pm$ 0.75& 71.18 $\pm$ 1.36& 68.37 $\pm$ 1.67& 74.24 $\pm$ 1.02\\
		& LACK & \textbf{90.48} $\pm$ \textbf{0.00} & \textbf{82.16} $\pm$ \textbf{0.00} & \textbf{81.66} $\pm$ \textbf{0.00} & \textbf{82.66} $\pm$ \textbf{0.00} \\ 
		\hline
		\hline
	\end{tabular}
\end{table}
\begin{table}[!t] \scriptsize
	\renewcommand{\arraystretch}{1.2}
	\caption{Average running times and speed up under different label ratios $\tau$.}
	\centering
	\label{table:C5}	
	\setlength{\tabcolsep}{1mm}
	\begin{tabular}{c|c|c|c|c|c|c|c}
		\hline
		\hline
		\multirow{2}{*}{$\tau$} & \multirow{2}{*}{Method} & \multicolumn{2}{|c}{Caltech-101-7}& \multicolumn{2}{|c}{Caltech-101-20}& \multicolumn{2}{|c}{MSRC-v1}\\ \cline{3-8} 
		&&Time&Speed up&Time&Speed up&Time&Speed up\\ \hline
		\multirow{3}{*}{$0$}& NMKC & 1.18 & 1.000$\times$ & 8.56 & 1.000$\times$  & 0.06 & 1.000$\times$\\
		& RMKMC & 6.06 & 0.195$\times$ & 25.89 & 0.331$\times$ & 0.26 & 0.231$\times$\\
		& LMVSC & 8.60 & 0.137$\times$ & 15.90 & 0.538$\times$  & 0.35 & 0.171$\times$\\\hline
		\multirow{6}{*}{$0.01$}& CK & 0.88 & 1.341$\times$ & 4.03 & 2.124$\times$  & \textbf{0.02} & \textbf{3.000$\times$}\\
		& MVCNMF & 41.29 & 0.029$\times$ & 204.17 & 0.042$\times$ & 0.20 & 0.300$\times$\\
		& MVOCNMF & 274.95 & 0.004$\times$ & 783.31 & 0.011$\times$ & 1.54 & 0.039$\times$\\
		& AMGL & 2.24 & 0.527$\times$ & 6.89 & 1.242$\times$  & \textbf{0.02} & \textbf{3.000$\times$}\\
		& MLAN & 4.45 & 0.265$\times$ & 23.92 & 0.358$\times$  & 0.04 & 1.500$\times$\\
		& LACK & \textbf{0.84} & \textbf{1.405$\times$} & \textbf{3.63} & \textbf{2.358$\times$}  & 0.05 & 1.200$\times$\\ \hline
		\multirow{6}{*}{$0.1$}& CK & 1.98 & 0.596$\times$ & 5.40 & 1.585$\times$  & \textbf{0.02} & \textbf{3.000$\times$}\\
		& MVCNMF & 33.79 & 0.035$\times$ & 154.68 & 0.055$\times$ & 0.17 & 0.353$\times$\\
		& MVOCNMF & 223.81 & 0.005$\times$ & 657.42 & 0.013$\times$ & 1.64 & 0.037$\times$\\
		& AMGL & 2.28 & 0.518$\times$ & 6.97 & 1.228$\times$  & \textbf{0.02} & \textbf{3.000$\times$}\\
		& MLAN & 5.31 & 0.222$\times$ & 22.44 & 0.382$\times$  & 0.07 & 0.857$\times$\\
		& LACK & \textbf{0.99} & \textbf{1.192$\times$} & \textbf{3.46} & \textbf{2.474$\times$}  & 0.03 & 2.000$\times$\\
		\hline
		\hline
	\end{tabular}
\end{table}

\subsection{The Weight Change Experiments}
To validate the effectiveness of our proposed label-driven auto-weighted strategy, we conduct classification experiments on the multi-view datasets with low-quality views. To simulate a multi-view dataset with low-quality views, we added a fake view to the Caltech-101-7 dataset and Gaussian noise to two original views on the Caltech-101-20 dataset. As the comparison method, we devised two methods based on different weighted strategies. One is the MLCK method that treats different views equally, and the other is the DACK method that adaptively adjusts the weight by the data-driven strategy Eq.\eqref{weight_RMVKM}. The only difference between MLCK, DACK, and our proposed LACK is a different strategy for the weight of views.

\subsubsection{Caltech-101-7 dataset with fake view}
\label{Fake_views}
\begin{figure}[!t]
	\centering
	\subfigure[DACK ($\tau=0.01$).]{
		\includegraphics[width=0.23\textwidth]{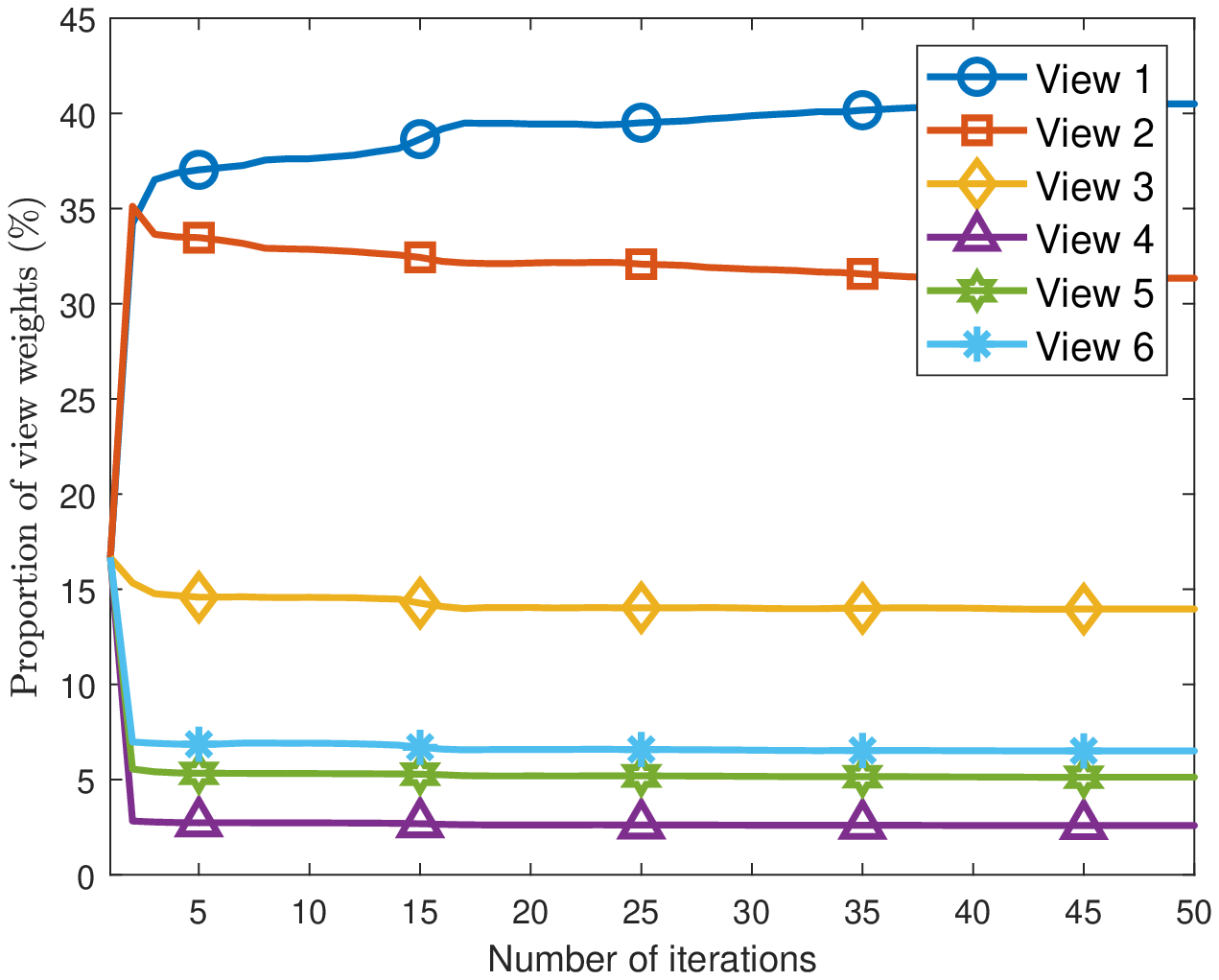}}
	\subfigure[DACK ($\tau=0.1$)]{
		\includegraphics[width=0.23\textwidth]{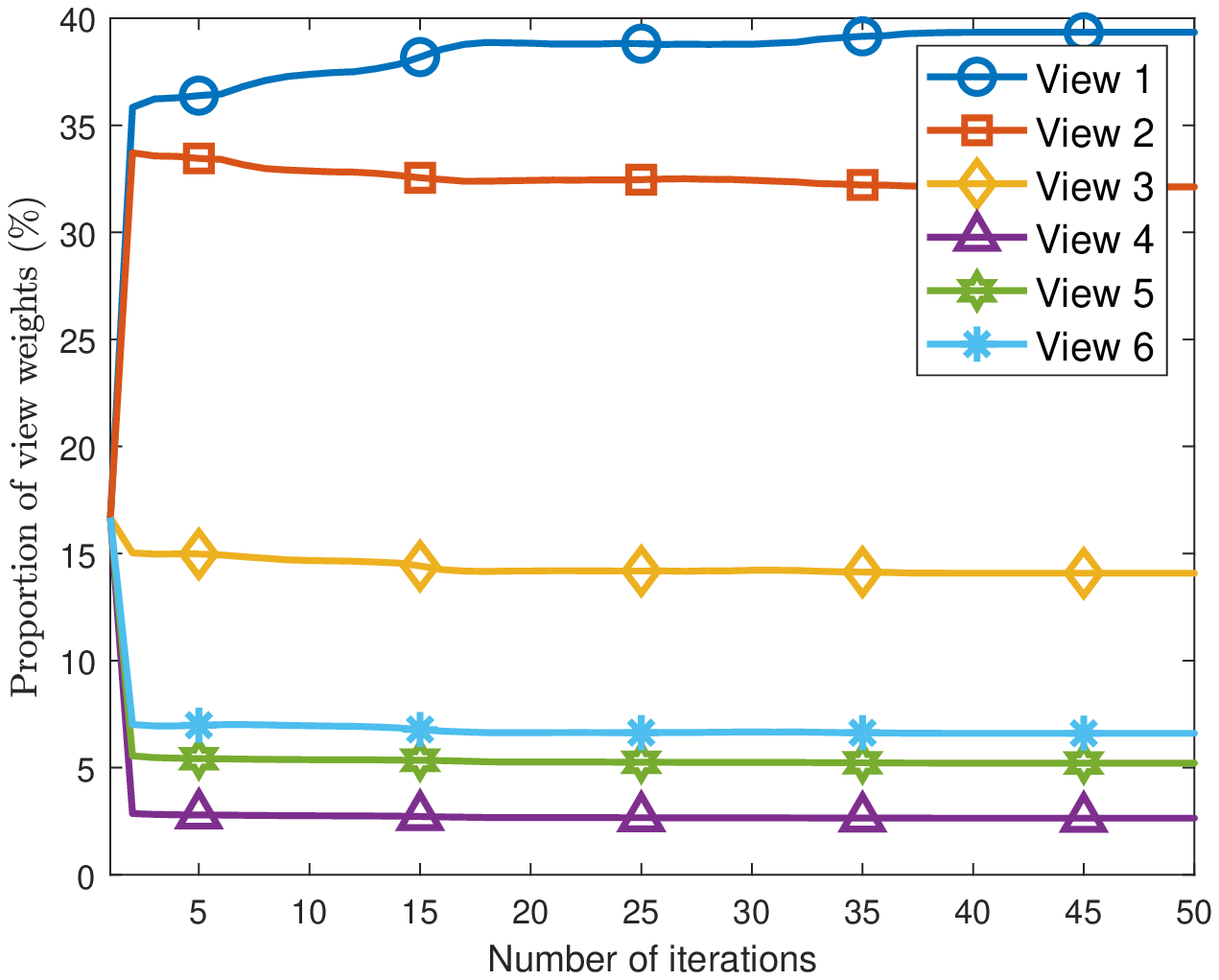}}\\
	\subfigure[LACK ($\tau=0.01$).]{
		\includegraphics[width=0.23\textwidth]{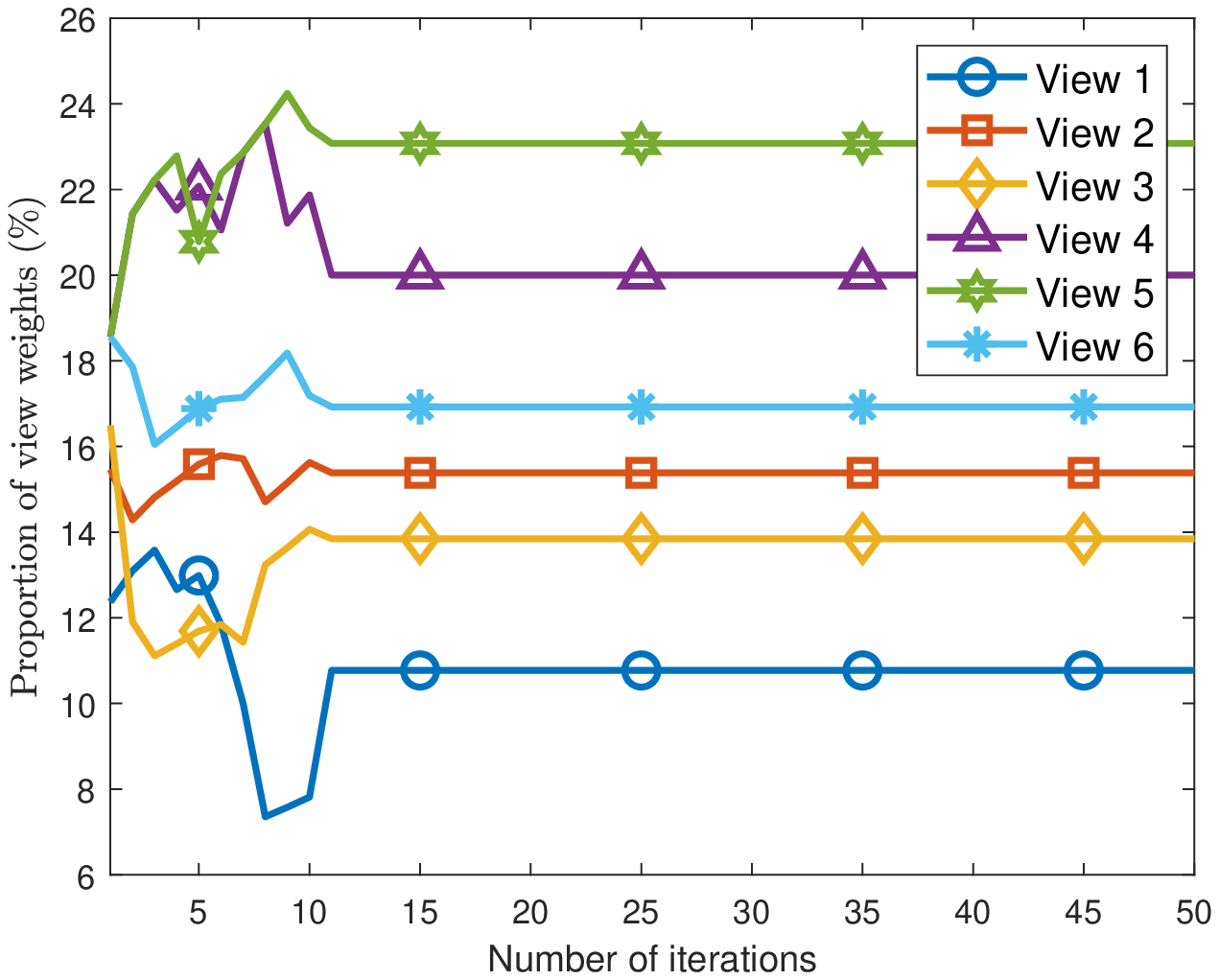}}
	\subfigure[LACK ($\tau=0.1$)]{
		\includegraphics[width=0.23\textwidth]{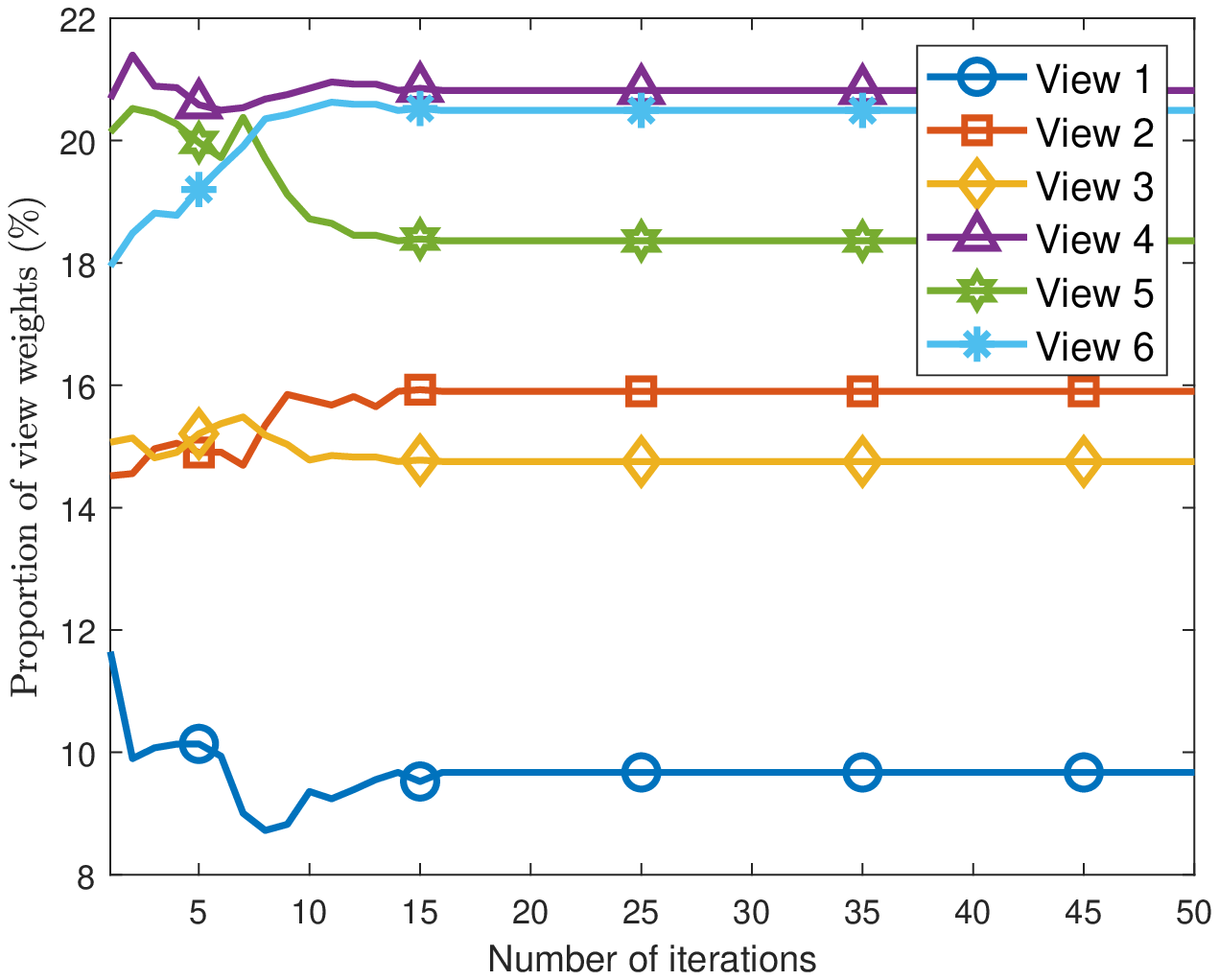}}\\
	\subfigure[DACK ($\tau=0.01$).]{
		\includegraphics[width=0.23\textwidth]{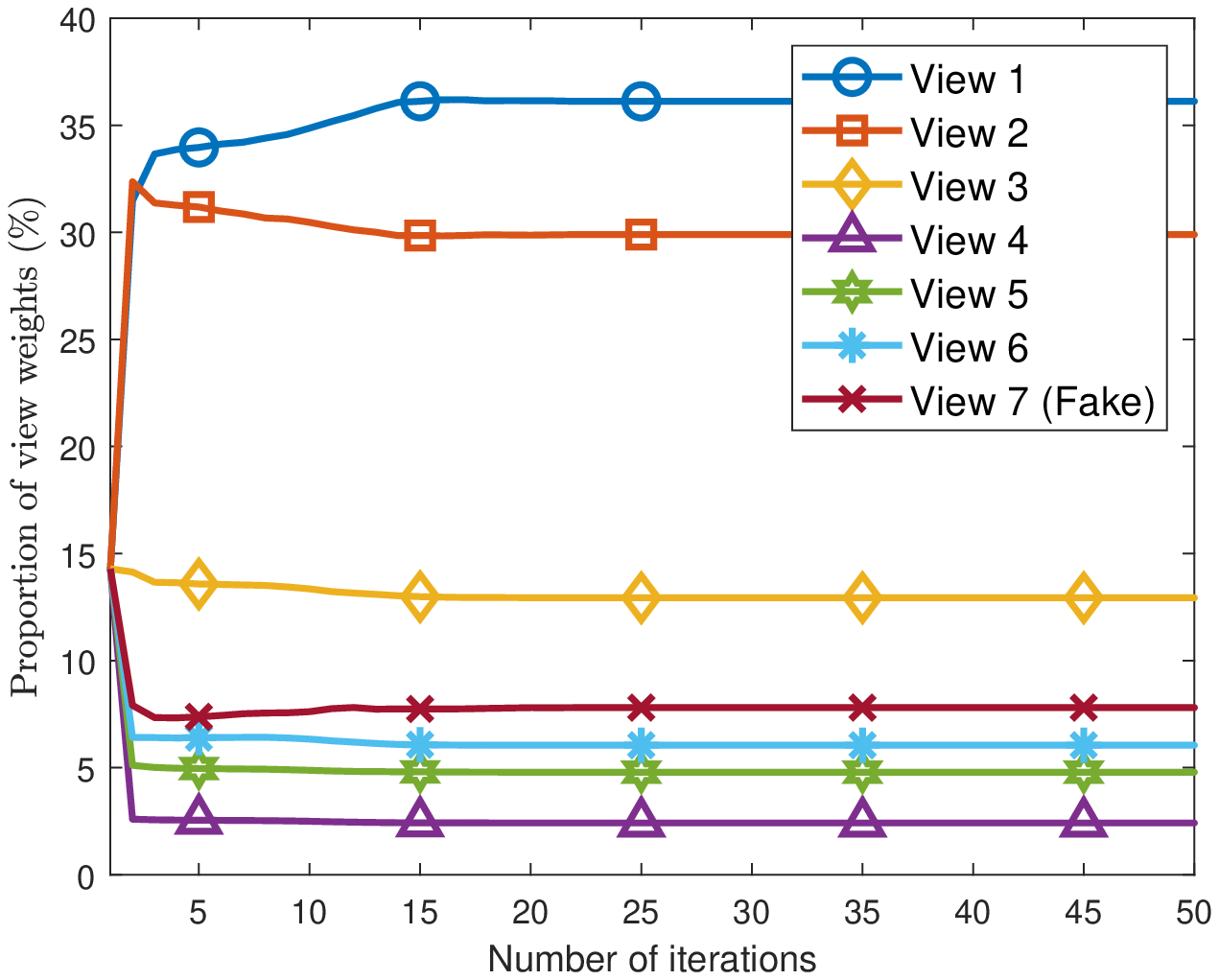}}
	\subfigure[DACK ($\tau=0.1$)]{
		\includegraphics[width=0.23\textwidth]{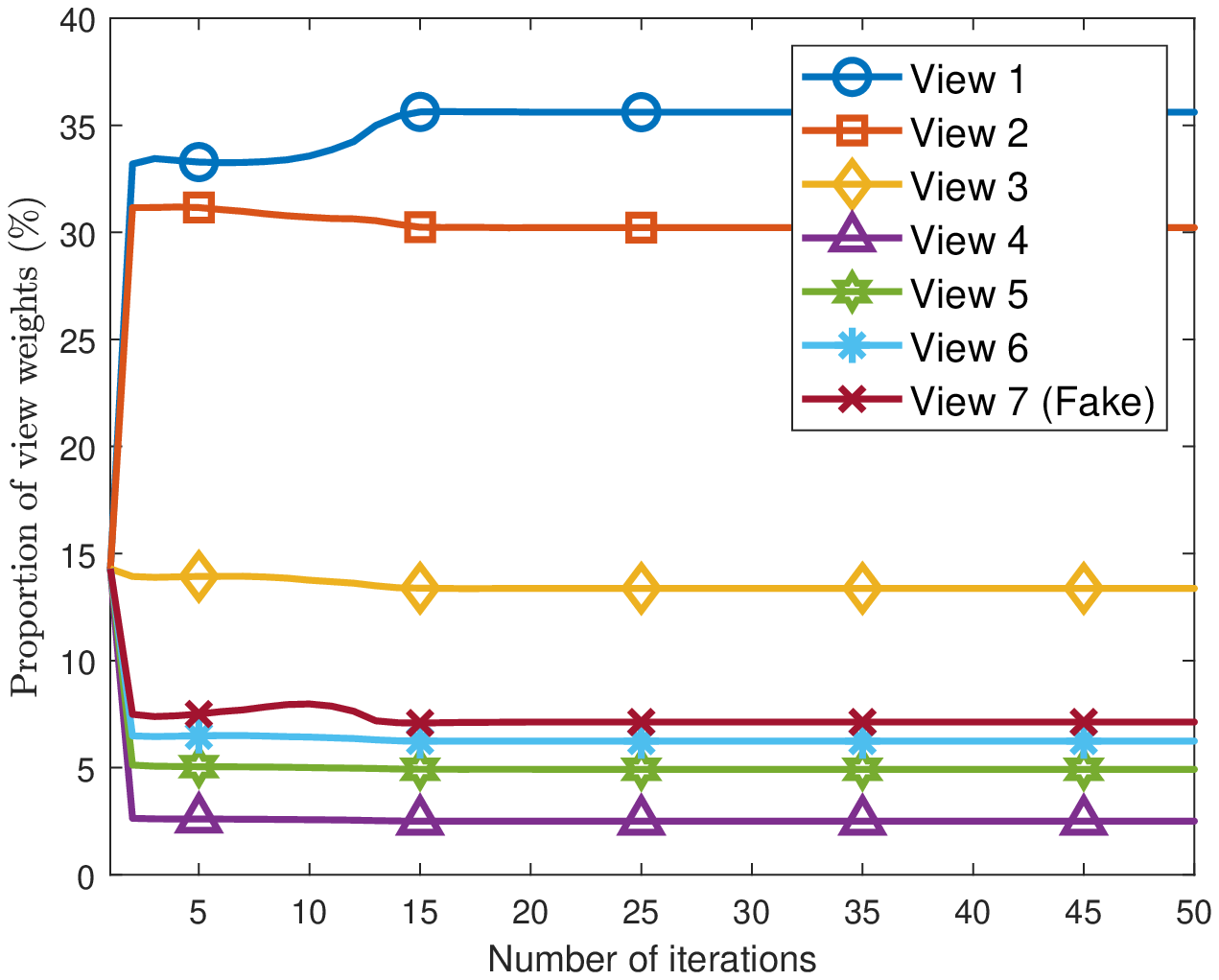}}\\
	\subfigure[LACK ($\tau=0.01$).]{
		\includegraphics[width=0.23\textwidth]{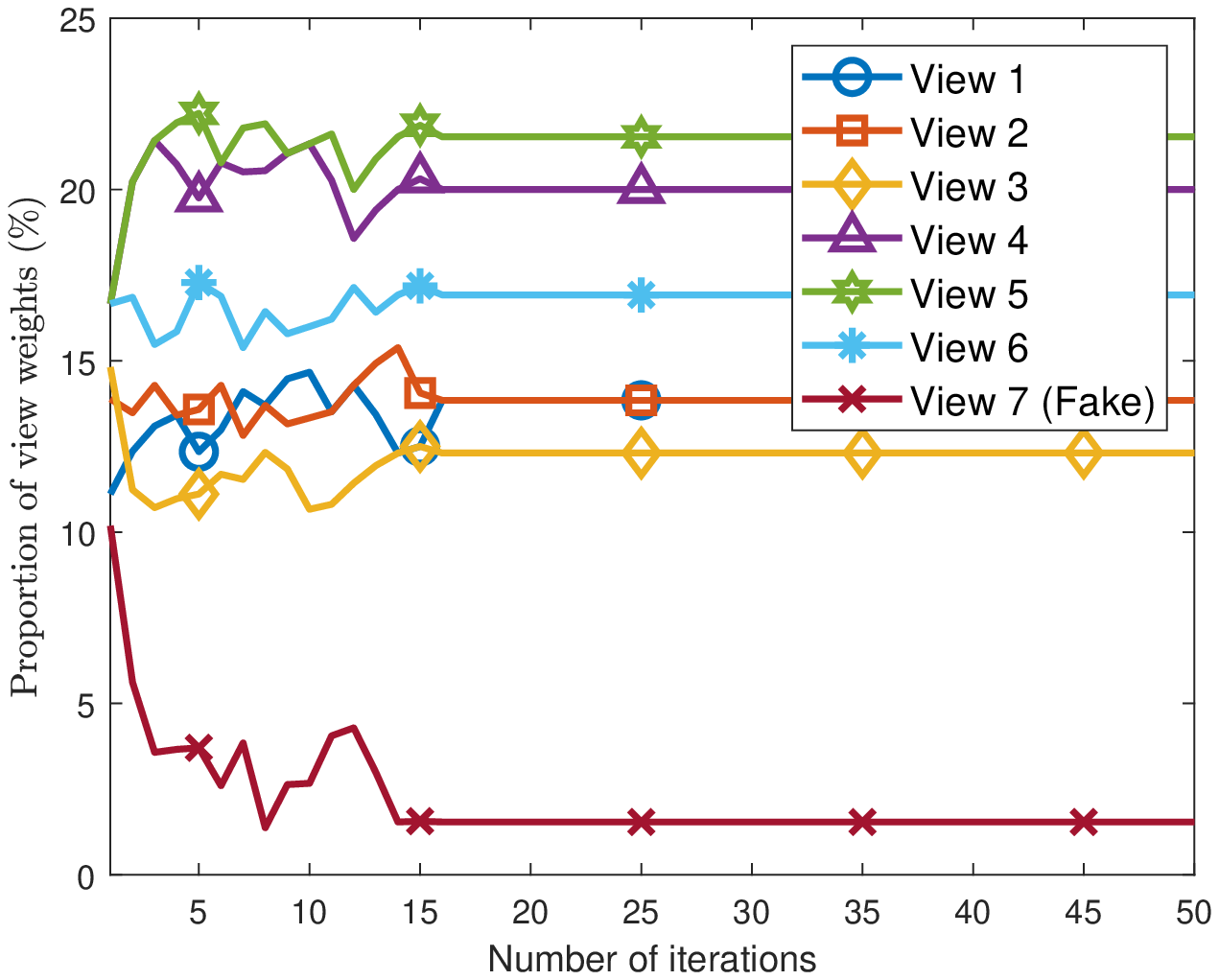}}
	\subfigure[LACK ($\tau=0.1$)]{
		\includegraphics[width=0.23\textwidth]{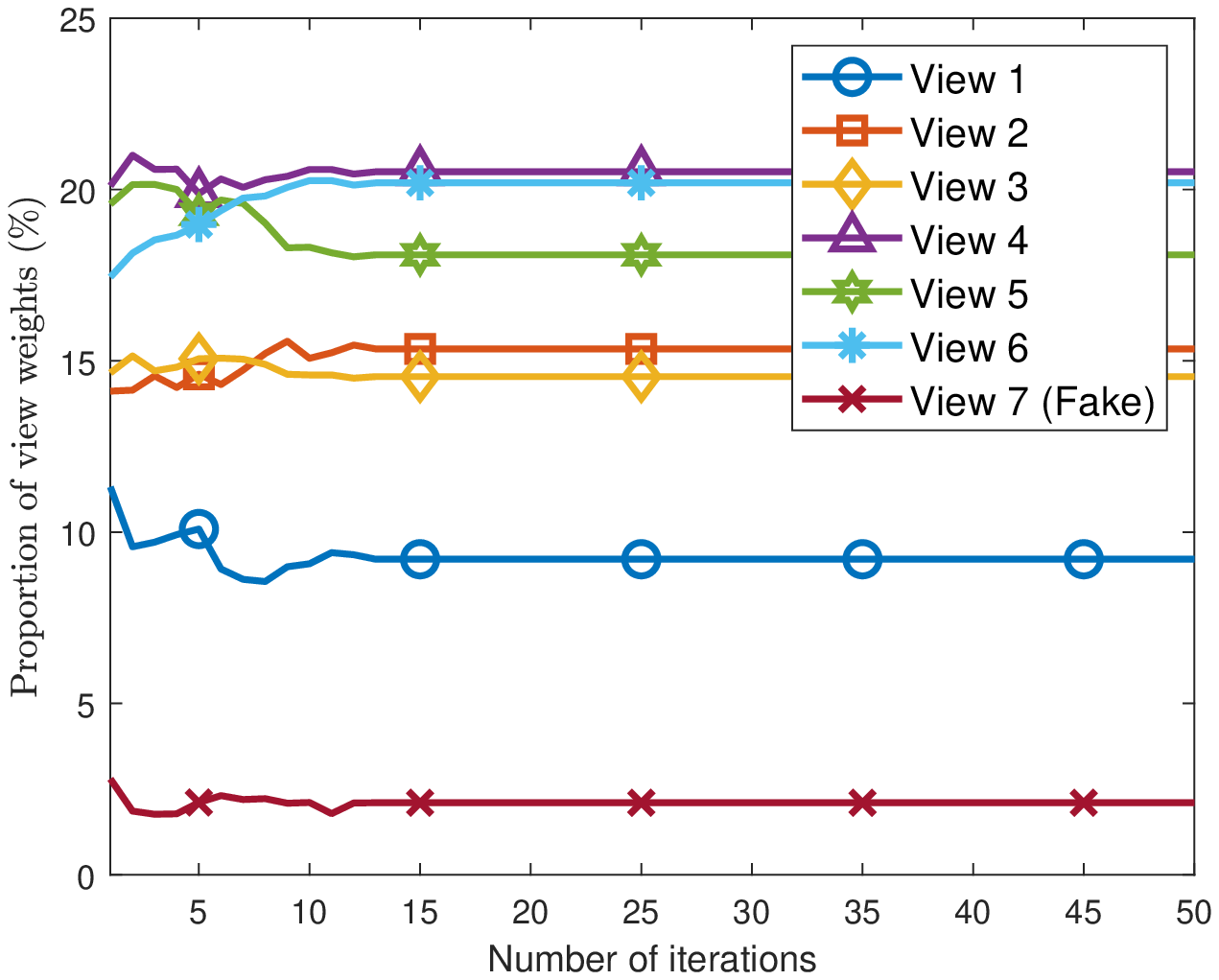}}
	\caption{The weight change curves of DACK and LACK on Caltech-101-7 dataset in pure and the low-quality(fake) view cases. The pure cases: (a)-(d); the low-quality (fake) view cases: (e)-(h).}
	\label{fig:fake}
\end{figure}

We report the clustering performance of K-means for each view of the Caltech-101-7 dataset in Fig. \ref{table:fake}. We can find that view 4 has the best clustering performance. It indicates that view $4$ is favorable to identify the dataset. We report the weight change curves of weight for DACK and LACK on the Caltech-101-20 dataset in Fig. \ref{fig:fake}. The results show that view 4 is secondary emphasized or emphasized in LACK while its importance is ignored in DACK when $\tau=\{0.01,0.1\}$, respectively, which indicates the rationality of the label-driven auto-weighted strategy.

To simulate the add low-quality view case, we construct two matrices $\mathbf{A} \in \mathbb{R}^{628 \times 7}$ and $\mathbf{B} \in \mathbb{R}^{7 \times 1474}$ using uniform distributed of random numbers in the interval $\{0,1\}$, and then construct a low-rank matrix $\mathbf{X}_7$ with elements of random values from $\mathbf{X}_7=\mathbf{A} \mathbf{B}$. The matrix $\mathbf{X}_7$ is called the fake view and is incorporated into the Caltech-101-7 dataset as $7$th view to simulate the low-quality dataset with the fake view. From Table \ref{table:fake}, we find that the fake view is not conducive to the clustering task, so the ideal auto-weighted strategy needs to adaptively reduce the weight of the fake view.

From Table \ref{table:fake} and Fig. \ref{fig:fake}, we can observe that the following: as the weight of each view of MLCK is equal, the classification performance of MLCK is severely degraded by the interference of fake views; DACK does not adaptively minimize the weight of fake views. The reason is that fake views are carefully faked and can deceive the model very well. Therefore, the data-driven strategy cannot significantly reduce the weight of fake view 7; LACK can minimize the weight of view 7 and provide better classification performance than the other methods. The reason is that LACK distinguishes view importance by evaluating the quality of the learned centroid matrix, and the quality of the centroid matrix for fake view 7 without clustering structure will obviously be much lower than the quality of the centroid matrix for other views, so the weight of fake view 7 is correctly minimized.

\begin{table}[!t] \scriptsize
	\renewcommand{\arraystretch}{1.2}
	\caption{The classification experiment of K-means, MLCK, DACK, and LACK under the different label ratios in the Caltech-101-7 dataset (pure and low-quality cases).}
	\centering
	\label{table:fake}
	\setlength{\tabcolsep}{1.6mm}
	\begin{tabular}{c|c|cccccc}
		\hline
		\hline
		Caltech-101-7&$\tau$ & Method & ACC & F-score & Precision & Recall & Time\\ \hline
		\multirow{12}{*}{Pure}& \multirow{6}{*}{$0$} &K-means(1) & 33.81 & 32.63 & 57.62 & 22.76 & 0.01\\
		&&K-means(2) & 44.55 & 51.01 & 72.00 & 39.49 & 0.01\\
		&&K-means(3) & 39.21 & 39.55 & 65.73 & 28.28 & 0.01\\
		&&K-means(4) & 49.94 & 54.78 & 87.36 & 39.90 & 0.09\\
		&&K-means(5) & 47.50 & 55.45 & 90.86 & 39.90 & 0.02\\
		&&K-means(6) & 48.09 & 48.83 & 78.96 & 35.35 & 0.04\\ \cline{2-8}
		&\multirow{3}{*}{$0.01$} & MLCK & 79.12 & 81.29 & 96.53 & 70.20 & 1.37\\
		&& DACK & 46.98 & 53.98 & 88.12 & 38.90 & 3.61 \\
		&& LACK & \textbf{82.01}  & \textbf{81.31}  & \textbf{94.81}  & \textbf{71.17}  & \textbf{0.84} \\ \cline{2-8}
		&\multirow{3}{*}{$0.1$}& MLCK & 81.48 & 81.70 & 97.86 & 70.12 & \textbf{0.82} \\
		&& DACK & 56.61 & 60.63 & 90.49 & 45.59& 2.93 \\
		&& LACK & \textbf{82.01} & \textbf{82.31}  & \textbf{97.91}  & \textbf{71.00} & 0.99 \\
		\hline
		\multirow{7}{*}{Low-quality}& 0 &K-means(7) & 20.34 & 22.25 & 38.44 & 15.66 & 0.05 \\ \cline{2-8}
		&\multirow{3}{*}{$0.01$}& MLCK & 63.53 & 62.43 & 91.11 & 47.48 & 2.49\\
		&& DACK & 48.97 & 52.55 & 85.70 & 37.89 & 2.13\\
		&& LACK & \textbf{79.26} & \textbf{81.57} & \textbf{96.91} & \textbf{70.43} & \textbf{1.53}\\ \cline{2-8}
		&\multirow{3}{*}{$0.1$}& MLCK & 65.08 & 66.26 & 93.33 & 51.37 & 2.05\\
		&& DACK & 57.60 & 61.31 & 90.76 & 46.29 & 1.81\\
		&& LACK & \textbf{81.94} & \textbf{82.28} & \textbf{97.84} & \textbf{71.00} & \textbf{0.94}\\
		\hline
		\hline
	\end{tabular}
\end{table}

\subsubsection{Caltech-101-20 dataset with noisy views}
\begin{figure}[!t]
	\centering  
	\subfigure[DACK ($\tau=0.01$).]{
		\includegraphics[width=0.23\textwidth]{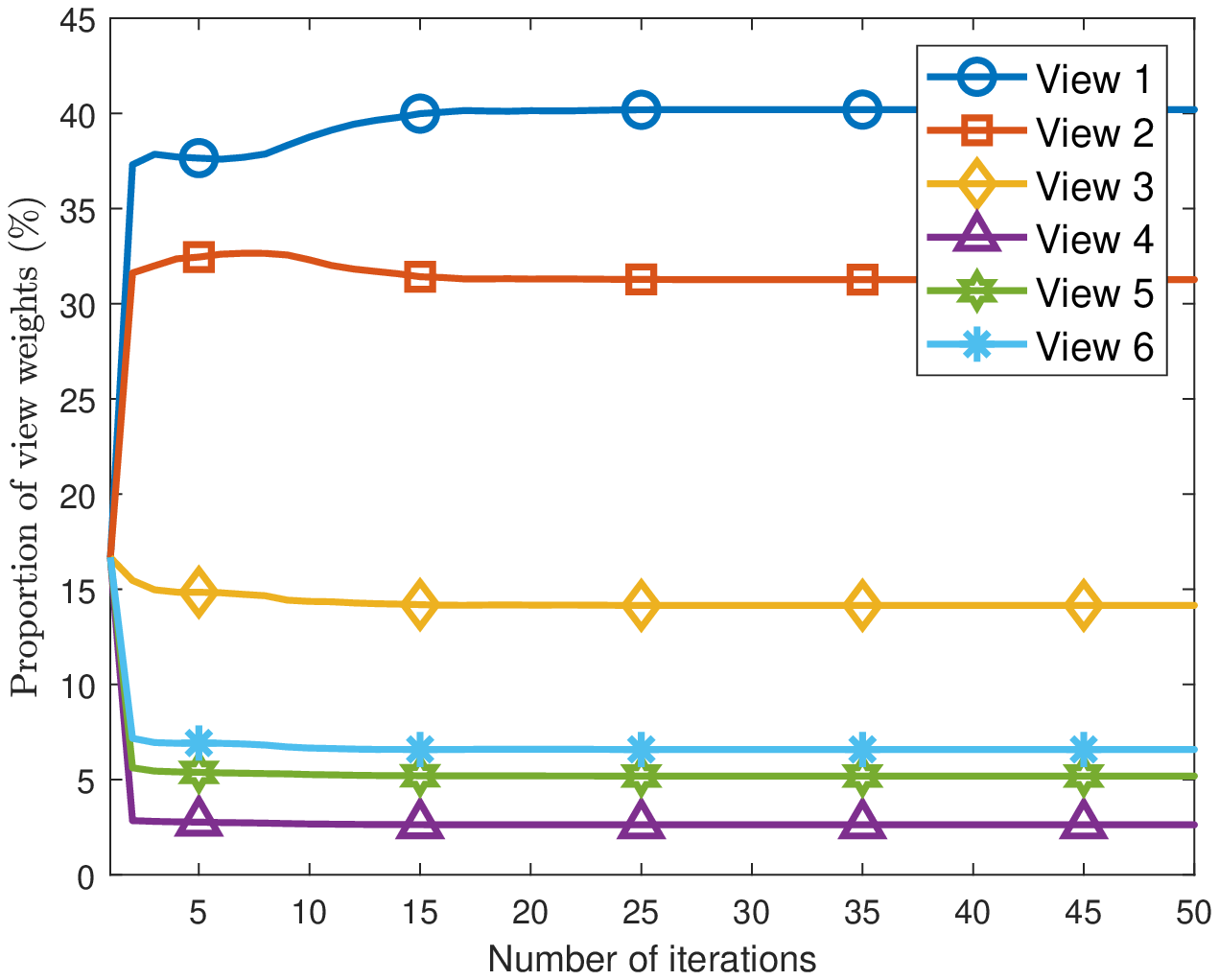}}
	\subfigure[DACK ($\tau=0.1$).]{
		\includegraphics[width=0.23\textwidth]{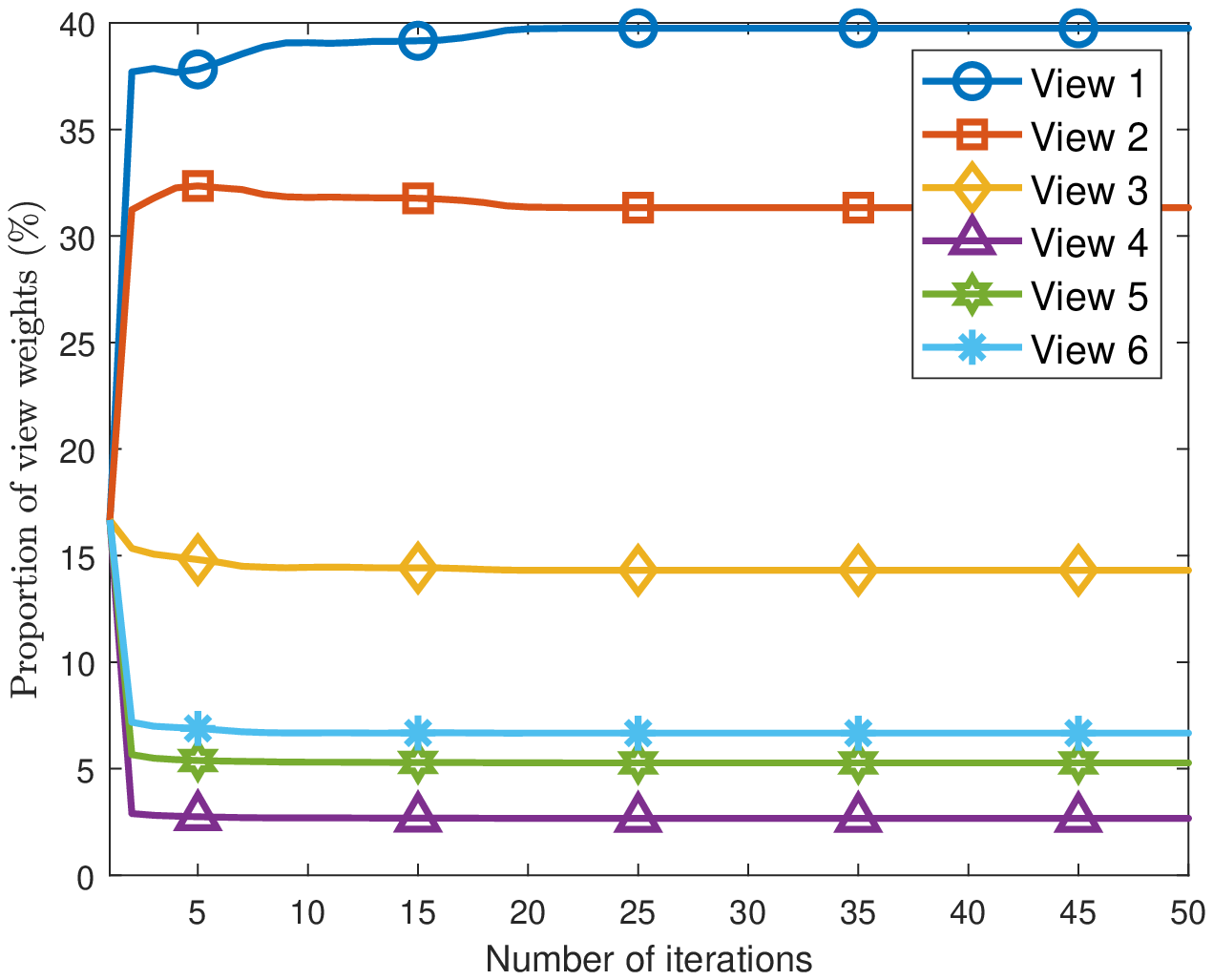}}\\
	\subfigure[LACK ($\tau=0.01$).]{
		\includegraphics[width=0.23\textwidth]{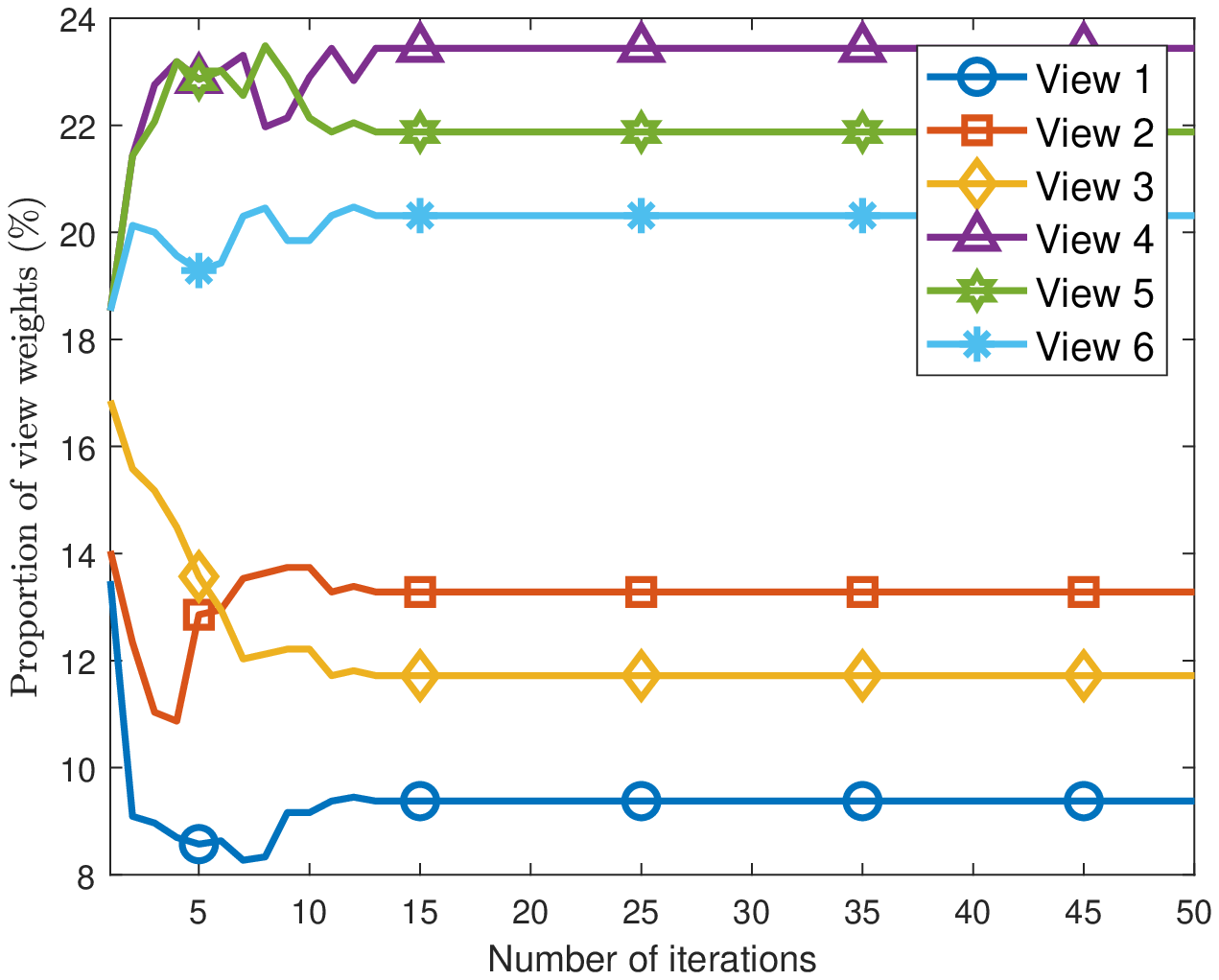}}
	\subfigure[LACK ($\tau=0.1$).]{
		\includegraphics[width=0.23\textwidth]{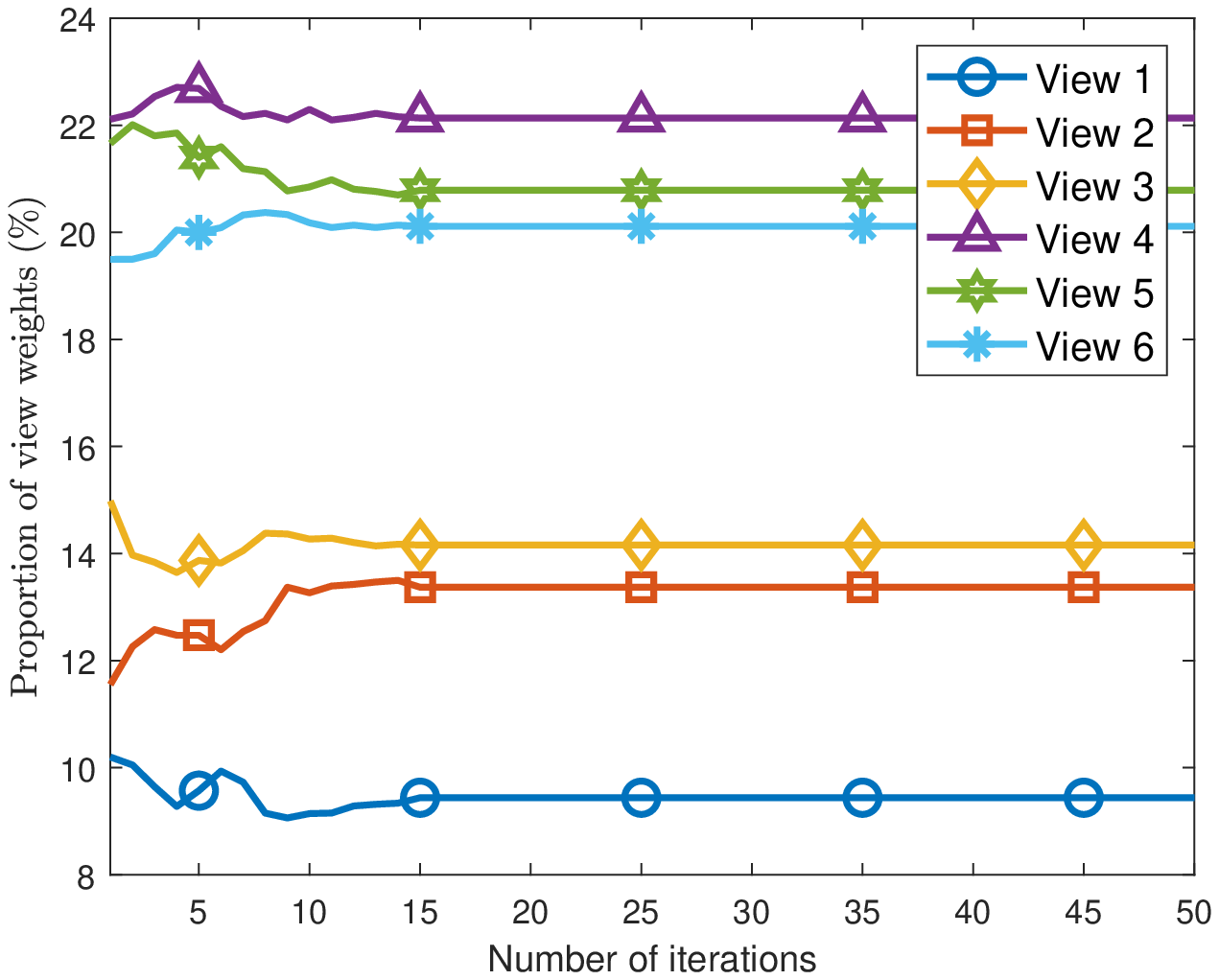}}\\
	\subfigure[DACK ($\tau=0.01$).]{
		\includegraphics[width=0.23\textwidth]{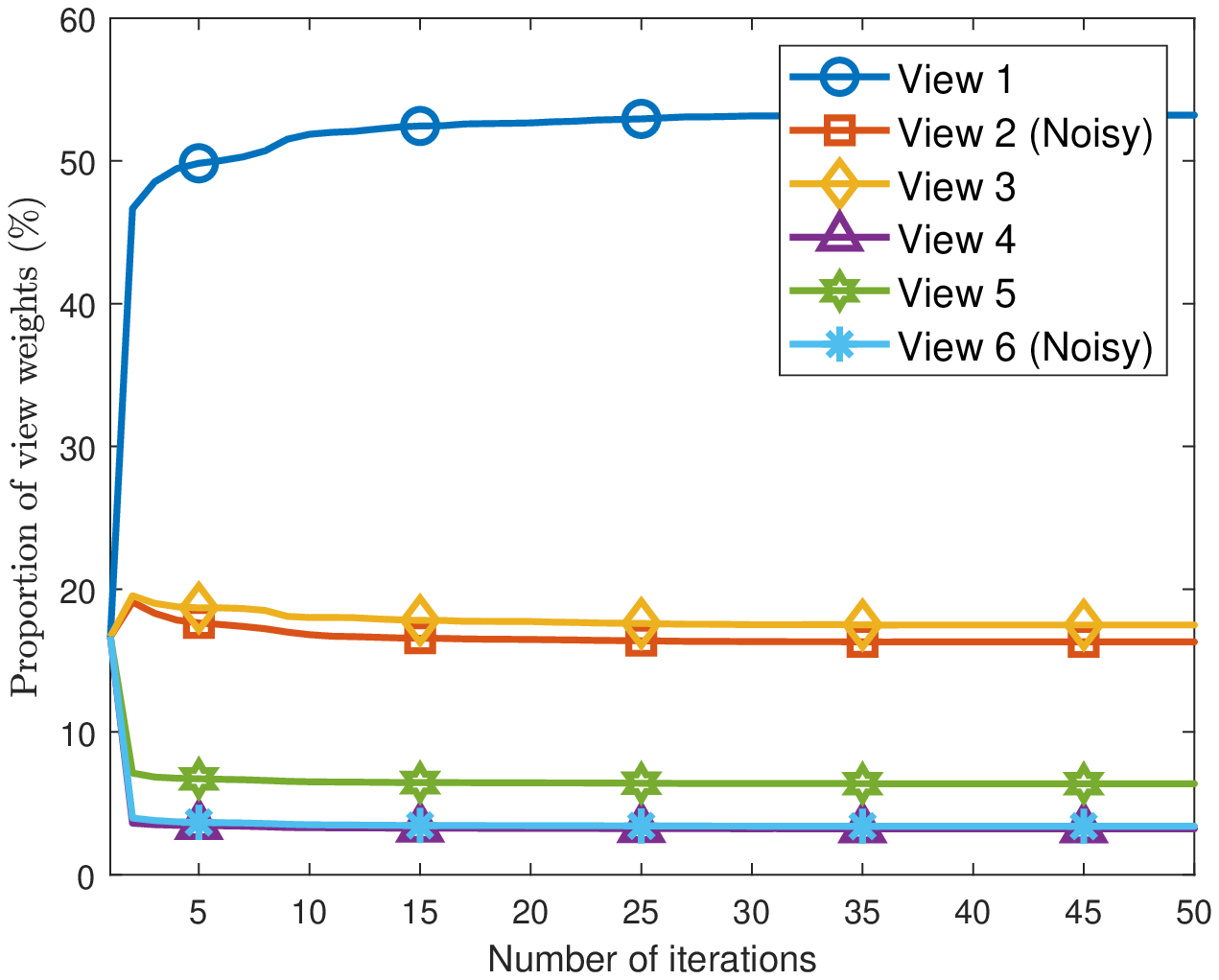}}
	\subfigure[DACK ($\tau=0.1$).]{
		\includegraphics[width=0.23\textwidth]{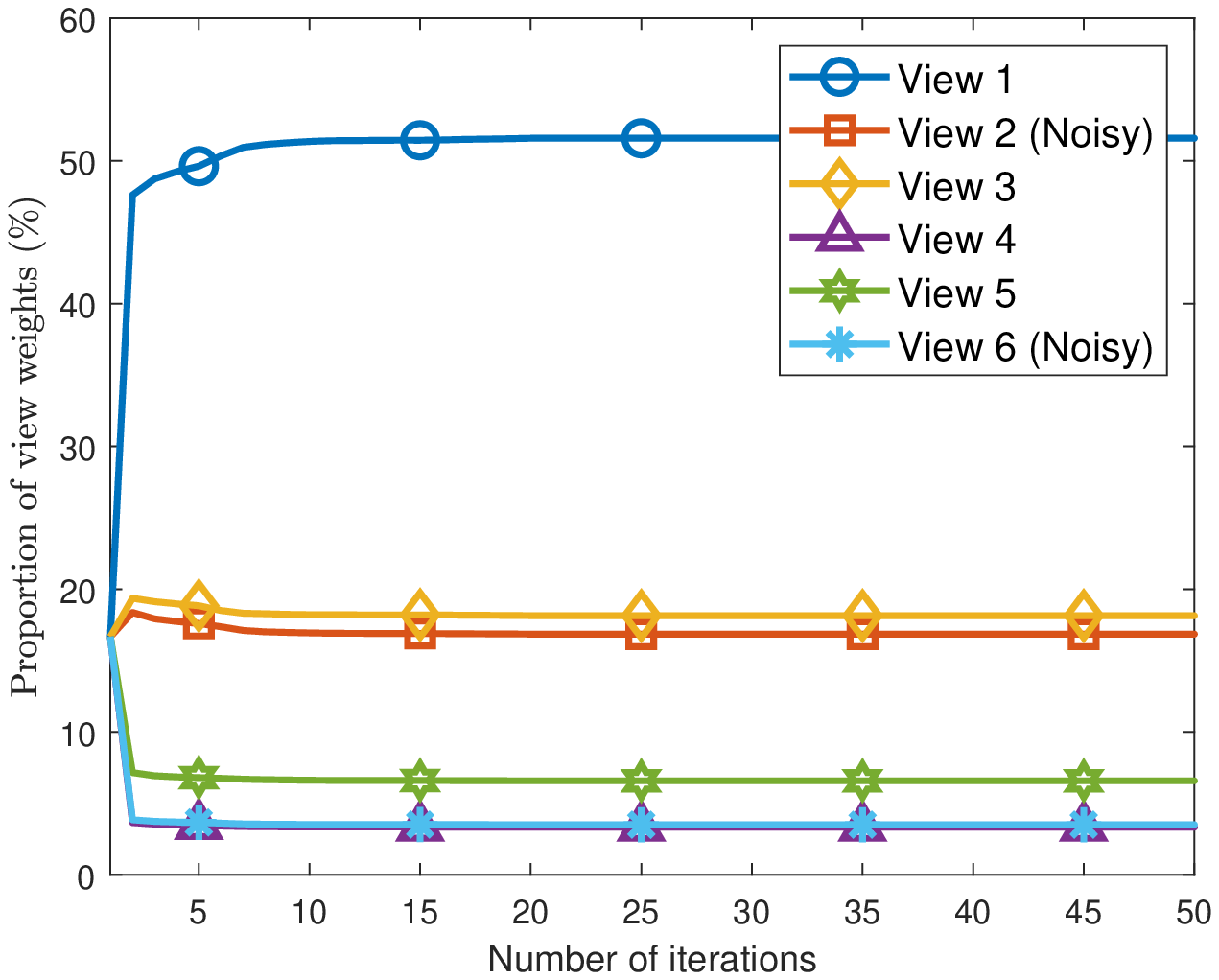}}\\
	\subfigure[LACK ($\tau=0.01$).]{
		\includegraphics[width=0.23\textwidth]{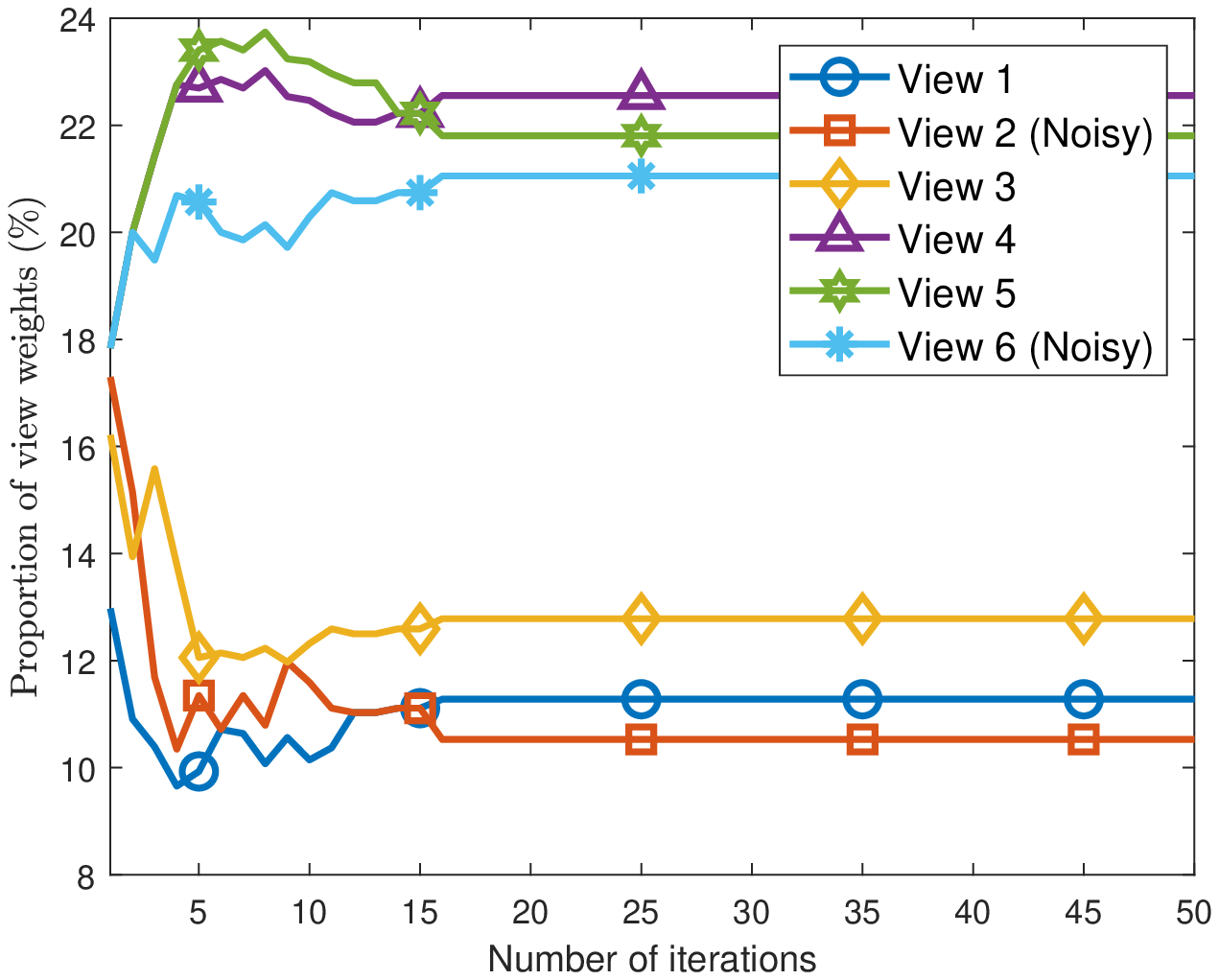}}
	\subfigure[LACK ($\tau=0.1$).]{
		\includegraphics[width=0.23\textwidth]{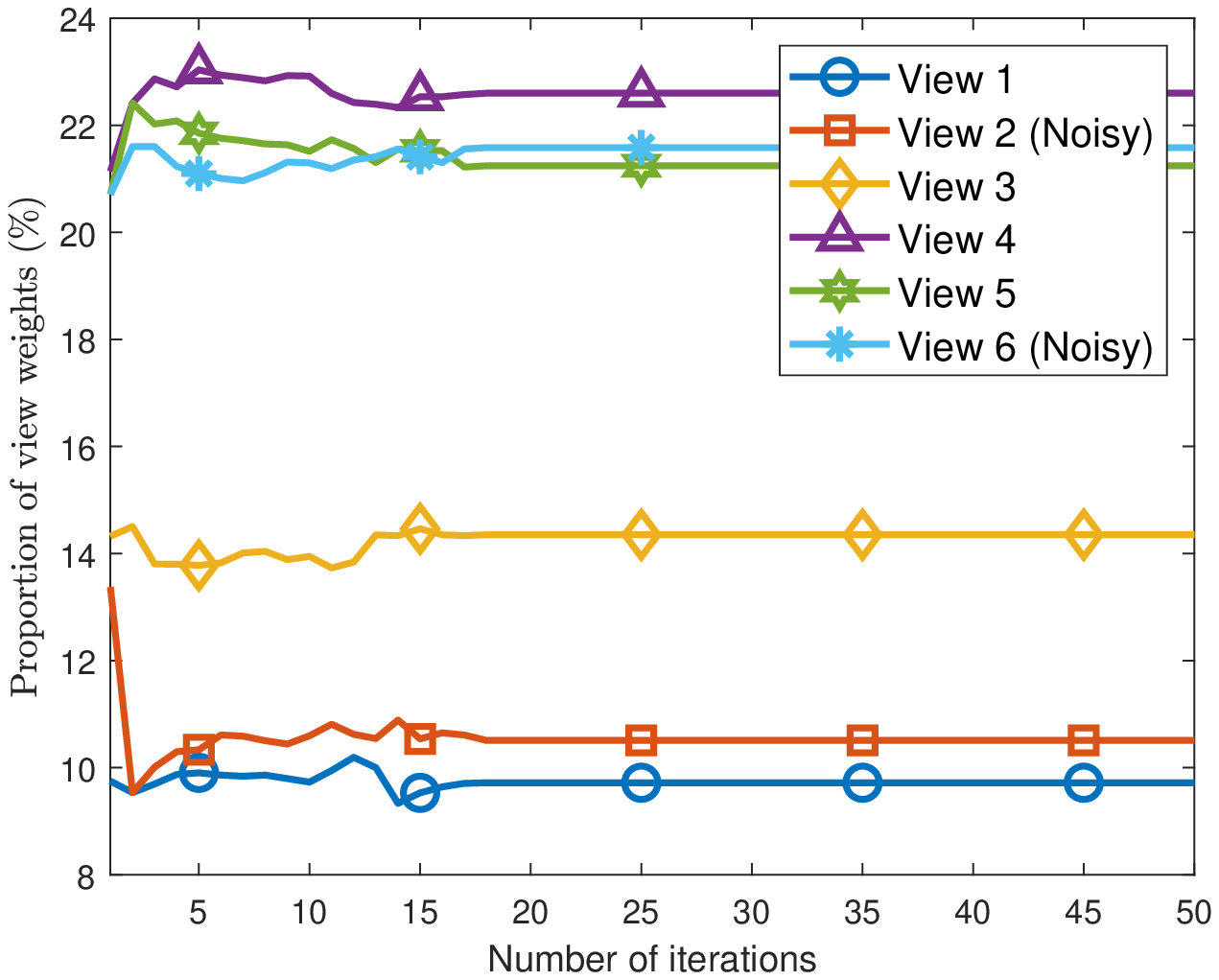}}
	\caption{The weight change curves of DACK and LACK on Caltech-101-20 dataset  in pure and low-quality (noisy) view cases.}
	\label{fig:noise}
\end{figure}

We report the clustering performance of K-means for each view of the Caltech-101-20 dataset in Table \ref{table:noise}. The clustering performance of views 4, 5, and 6 are the first three, which indicates that three views are more favorable to identify the dataset. We report the weight change curves of weight for DACK and LACK on the Caltech-101-20 dataset in Fig. \ref{fig:noise}. The results show that views $4$, $5$, and $6$ are emphasized in LACK, which indicates the rationality of the label-driven auto-weighted strategy.

To be fair, we choose the view set $\{2,6\}$ to simulate the low-quality views with noise because the clustering performance of views $\{2,6\}$ is moderate and relatively close. We have added the Gaussian noise with a signal-to-noise ratio of $1$ on the view set $\{2,6\}$. We merge the low-quality view set $\{2,6\}$ with the pure view set $\{1,3,4,5\}$ of Caltech-101-20 and call it Caltech-101-20 dataset with low-quality views. From Table \ref{table:noise}, it can observe that K-means has poor clustering performance on view 2, but better clustering performance on view 6. Therefore, the ideal multi-view method needs to adaptively reduce the weight of view 2 and give a high weight to view 6.

From the Table \ref{table:fake} and Fig. \ref{fig:fake}, we have: the classification performance of MLCK drops sharply due to equal treatment of low-quality views; since low-quality views are generally difficult to fit, DACK greatly reduces the weights of low-quality views 2 and 6 as shown in Fig. \ref{fig:fake}. However, the DACK greatly reduces the weight of view 6, which indicate that the data-driven auto-weighted strategy has the limitations to distinguish the importance of view; LACK greatly reduces the weight of view 2 and keeps the weight of view 6 almost constant, and then provides better classification performance, which is demonstrated that the effectiveness of our proposed strategy.

\begin{table}[!t] \scriptsize
	\renewcommand{\arraystretch}{1.2}
	\caption{The classification experiment of K-means, MLCK, DACK, and LACK under the different label ratios in the Caltech-101-20 dataset (pure and low-quality cases).}
	\label{table:noise}
	\centering
	\setlength{\tabcolsep}{1.6mm}
	\begin{tabular}{c|c|c|ccccc}
		\hline
		\hline
		Caltech-101-20&$\tau$ & Method & ACC & F-score & Precision & Recall & Time\\ \hline
		\multirow{10}{*}{Pure}& \multirow{6}{*}{$0$} &K-means(1) & 28.16 & 21.81 & 42.49 & 14.68 & 0.02\\
		&&K-means(2) & 33.11 & 30.16 & 47.39 & 22.18 & 0.02\\
		&&K-means(3) & 31.31 & 24.02 & 44.81 & 16.41 & 0.04\\
		&&K-means(4) & 49.46 & 44.04 & 76.31 & 30.98 & 0.19\\
		&&K-means(5) & 43.75 & 40.25 & 74.50 & 27.58 & 0.06\\
		&&K-means(6) & 37.33 & 34.01 & 61.24 & 23.54 & 0.11\\ \cline{2-8}
		&\multirow{3}{*}{$0.01$}& MLCK & \textbf{75.86} & 77.33 & 90.28 & 67.64 & 3.94 \\
		&& DACK & 52.74 & 53.24 & 78.81 & 40.19 & 7.36\\
		&& LACK & 74.50 & \textbf{77.37} & \textbf{90.34}  & \textbf{67.67} & \textbf{3.63} \\ \cline{2-8}
		&\multirow{3}{*}{$0.1$}& MLCK & 75.79 & 76.01 & \textbf{91.52} & 64.99 & 5.67 \\
		&& DACK & 55.84 & 54.64  & 80.99  & 41.23  & 5.46 \\
		&& LACK & \textbf{76.82}  & \textbf{77.59}  & 90.69  & \textbf{67.79} & \textbf{3.46} \\
		\hline
		\multirow{8}{*}{Low-quality}& \multirow{2}{*}{$0$}&K-means(2) & 21.91 & 17.59 & 35.49 & 11.69 & 0.02\\
		&&K-means(6) & 36.22 & 34.22 & 58.70 & 24.20 & 0.08\\ \cline{2-8}
		&\multirow{3}{*}{$0.01$}& MLCK & 66.34 & 69.79 & 85.71 & 58.85 & 6.72\\
		&& DACK & 50.02 & 52.34 & 72.90 & 40.82 & 10.07\\
		&& LACK & \textbf{72.21} & \textbf{78.08} & \textbf{89.06} & \textbf{69.52} & \textbf{3.98}\\ \cline{2-8}
		&\multirow{3}{*}{$0.1$}& MLCK & 69.91 & 67.90 & 86.05 & 56.07 & 4.69\\
		&& DACK & 57.48 & 57.21 & 78.86 & 44.88 & 4.99\\
		&& LACK & \textbf{78.22} & \textbf{78.35} & \textbf{90.76} & \textbf{68.93} & \textbf{4.17}\\
		\hline
		\hline
	\end{tabular}
\end{table}	

\subsection{The Stability Experiment}
To explore the dependence of the proposed label-driven auto-weighted strategy on the training data selection, we designed an experiment in this subsection. Firstly, on the Caltech-101-7 and Caltech-101-20 datasets, we randomly selected five different sets of training data and test data. Then experiments are conducted on these data using the LACK, MCCK, and DACK methods, respectively. Finally, the average performance of the methods is calculated and analyzed.

\begin{figure*}[!t]
	\centering
	\subfigure[ACC]{
		\includegraphics[width=0.23\textwidth]{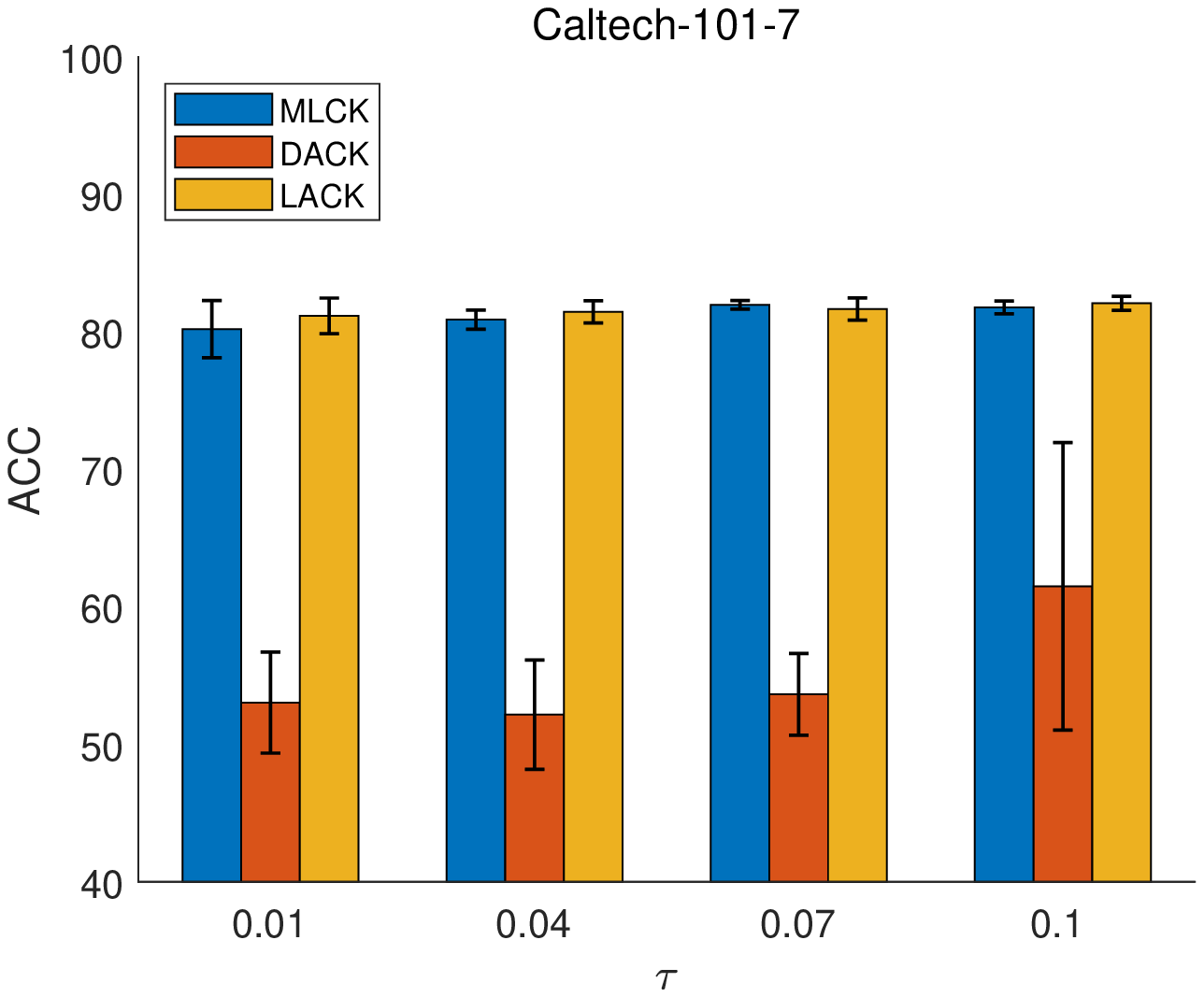}}
	\subfigure[F-score]{
		\includegraphics[width=0.23\textwidth]{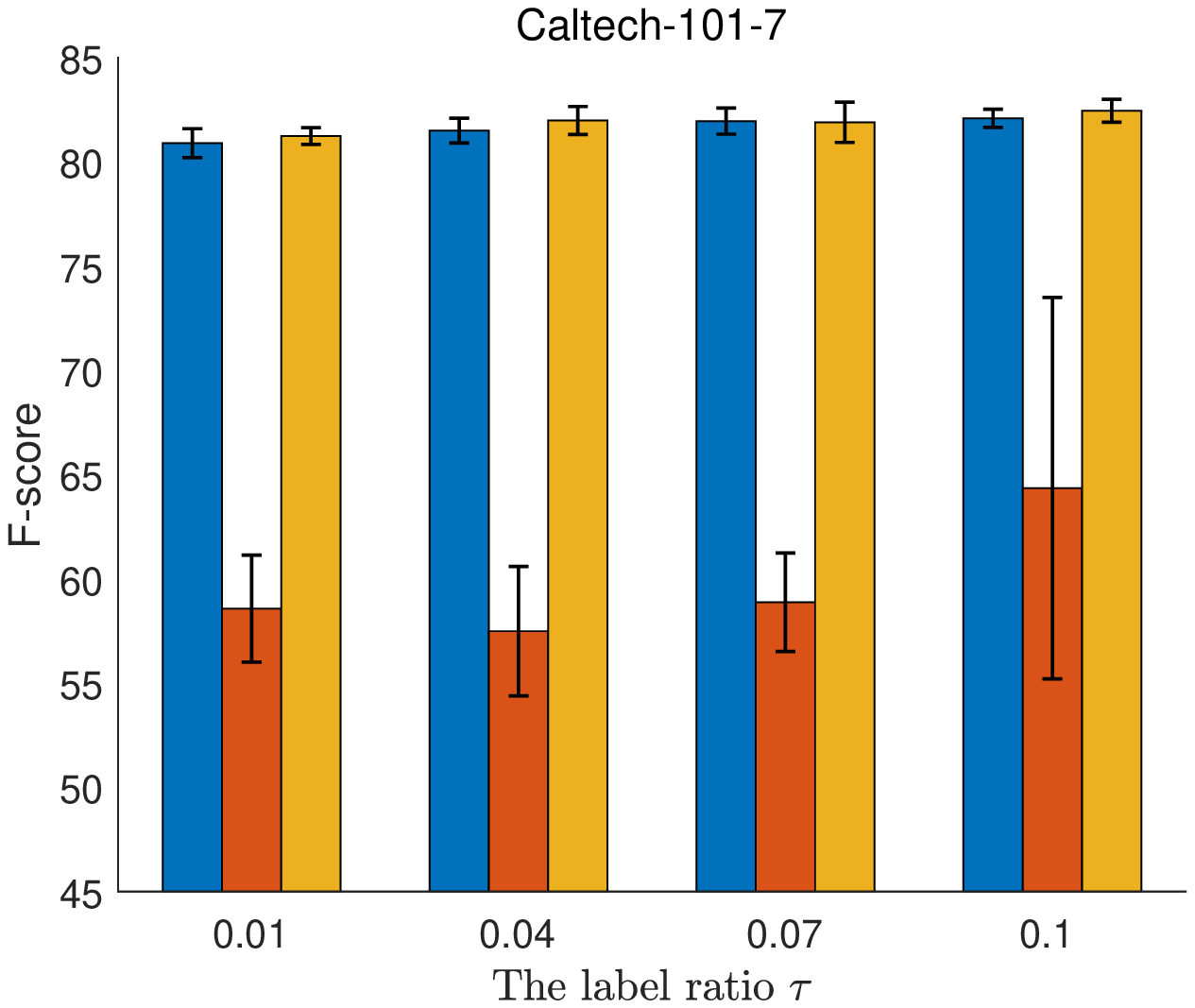}}
	\subfigure[Precision]{
		\includegraphics[width=0.23\textwidth]{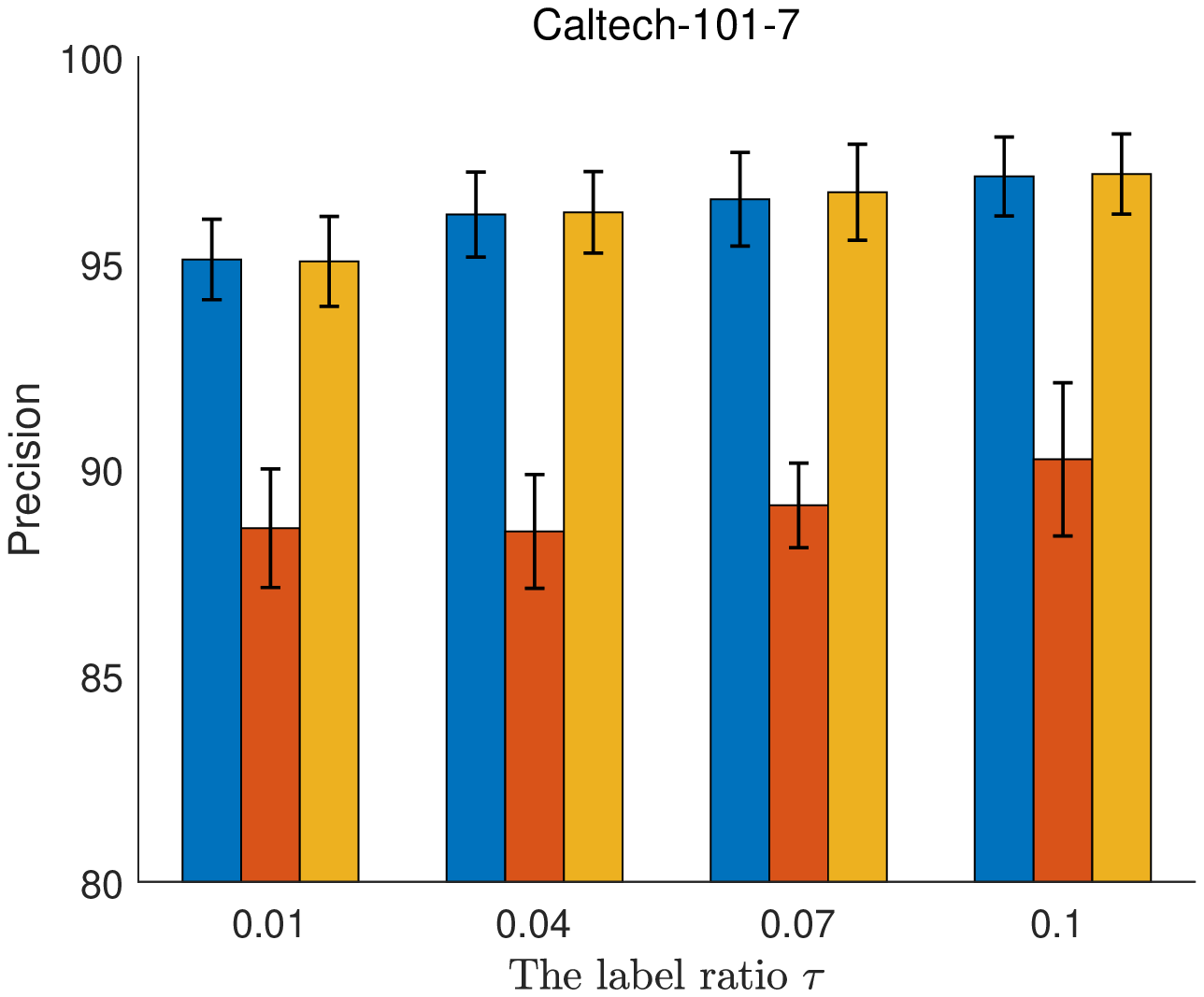}}
	\subfigure[Recall]{
		\includegraphics[width=0.23\textwidth]{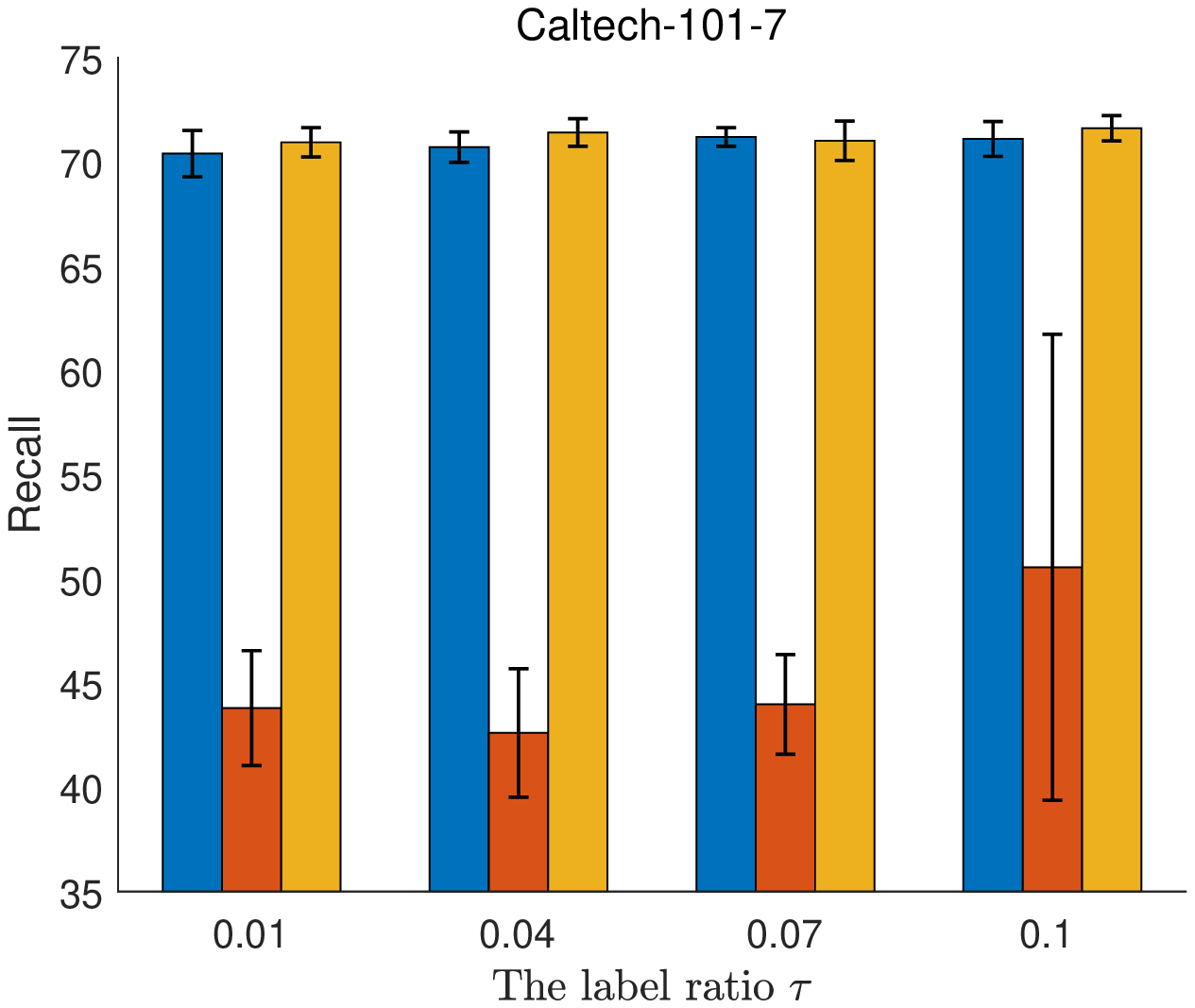}}
	\caption{The effect of the randomly selected training set in Caltech-101-7 dataset on the classification performance for the MLCK, DACK, and LACK methods.}
	\label{ranlabel_Cal-7_performance}
\end{figure*}

\begin{figure*}[!t]
	\centering
	\subfigure[ACC.]{
		\includegraphics[width=0.23\textwidth]{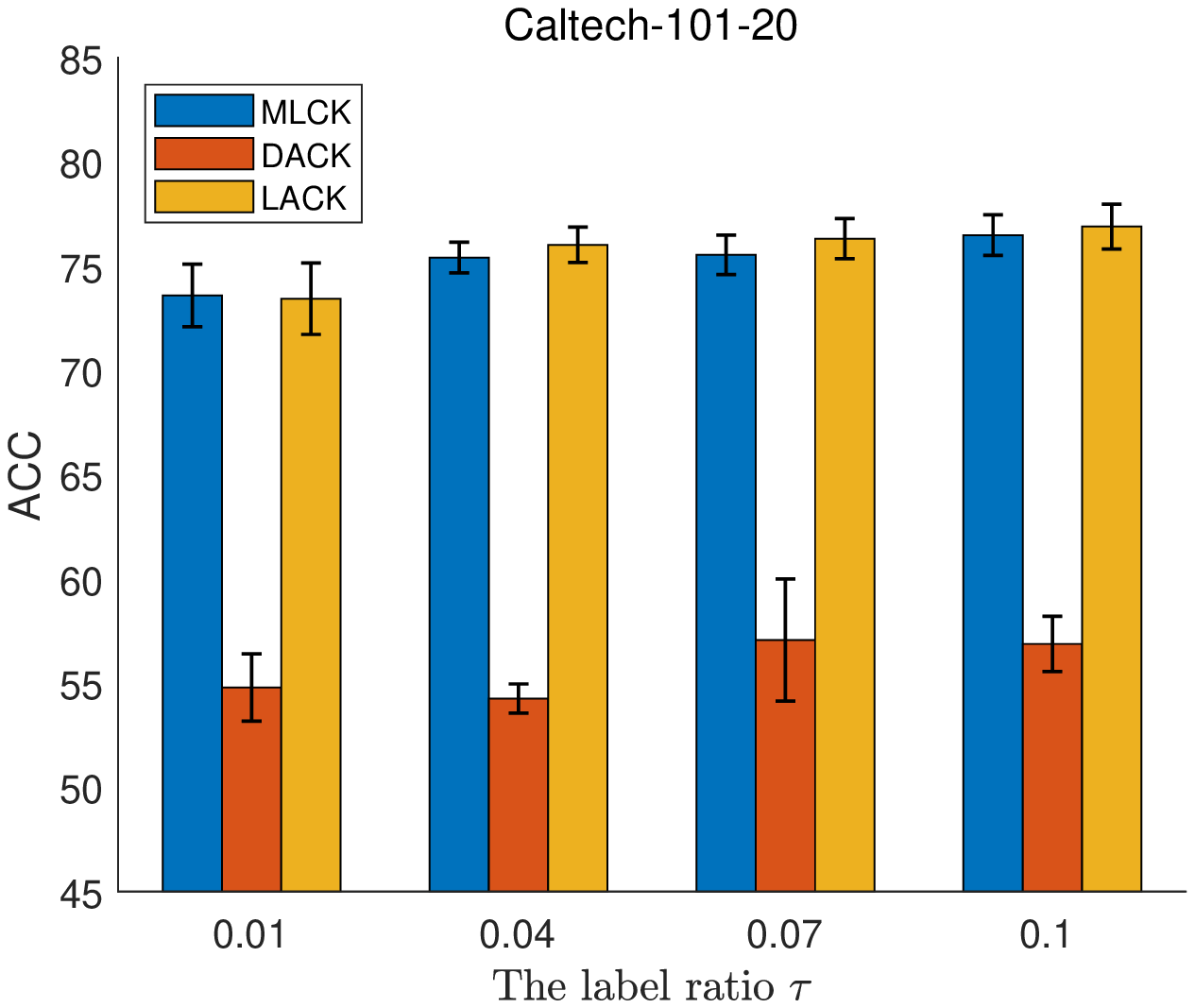}}
	\subfigure[F-score.]{
		\includegraphics[width=0.23\textwidth]{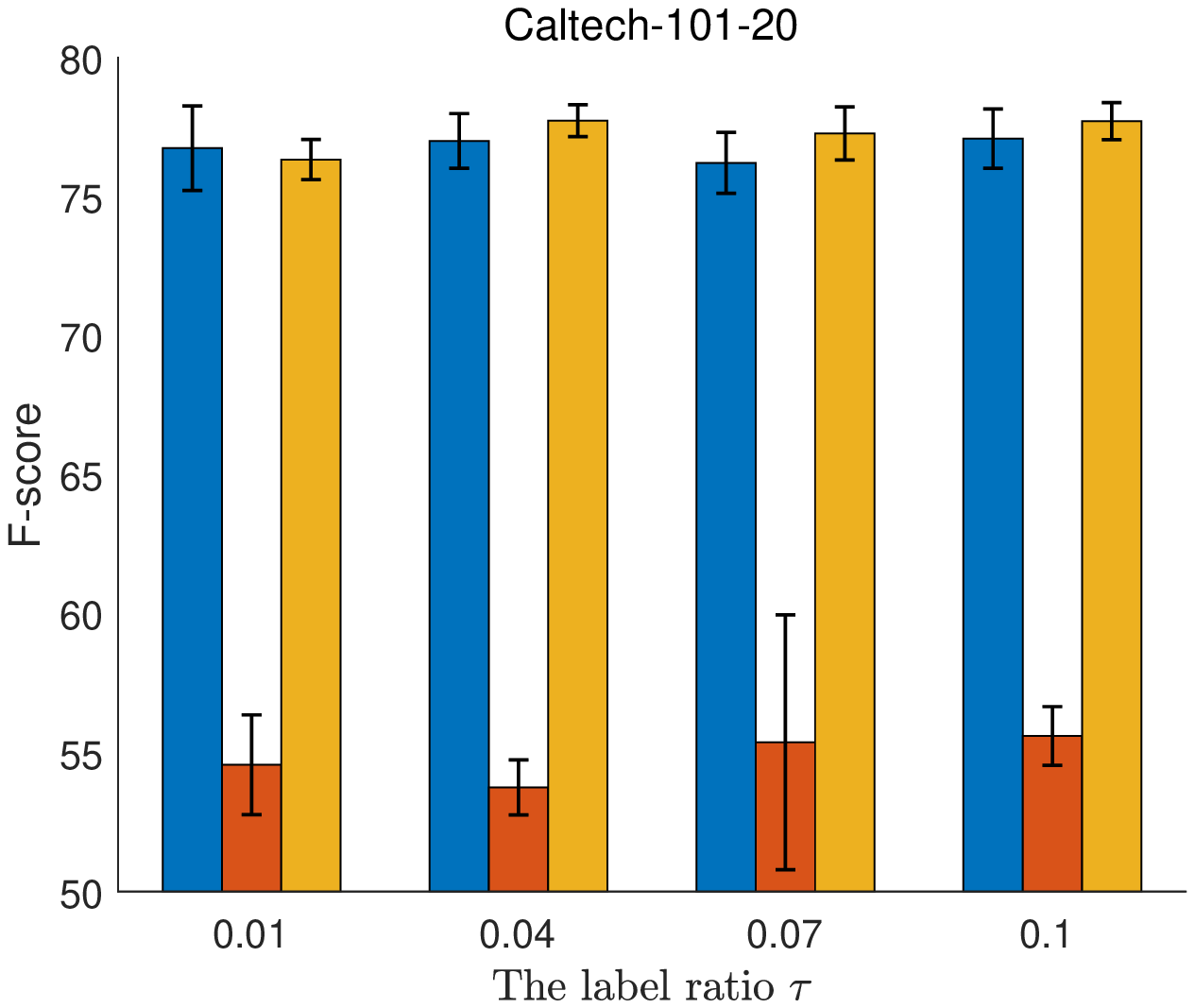}}
	\subfigure[Precision.]{
		\includegraphics[width=0.23\textwidth]{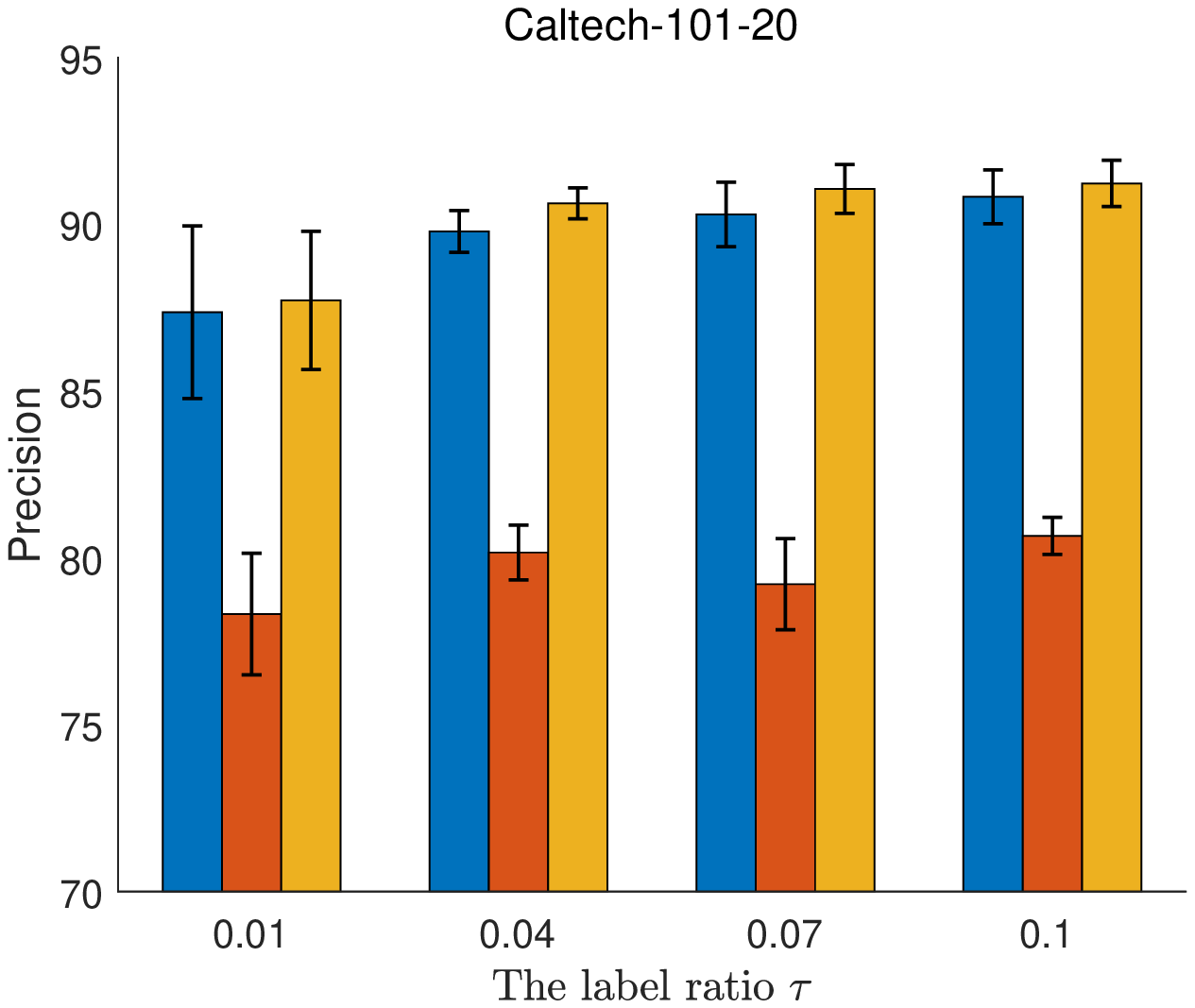}}
	\subfigure[Recall.]{
		\includegraphics[width=0.23\textwidth]{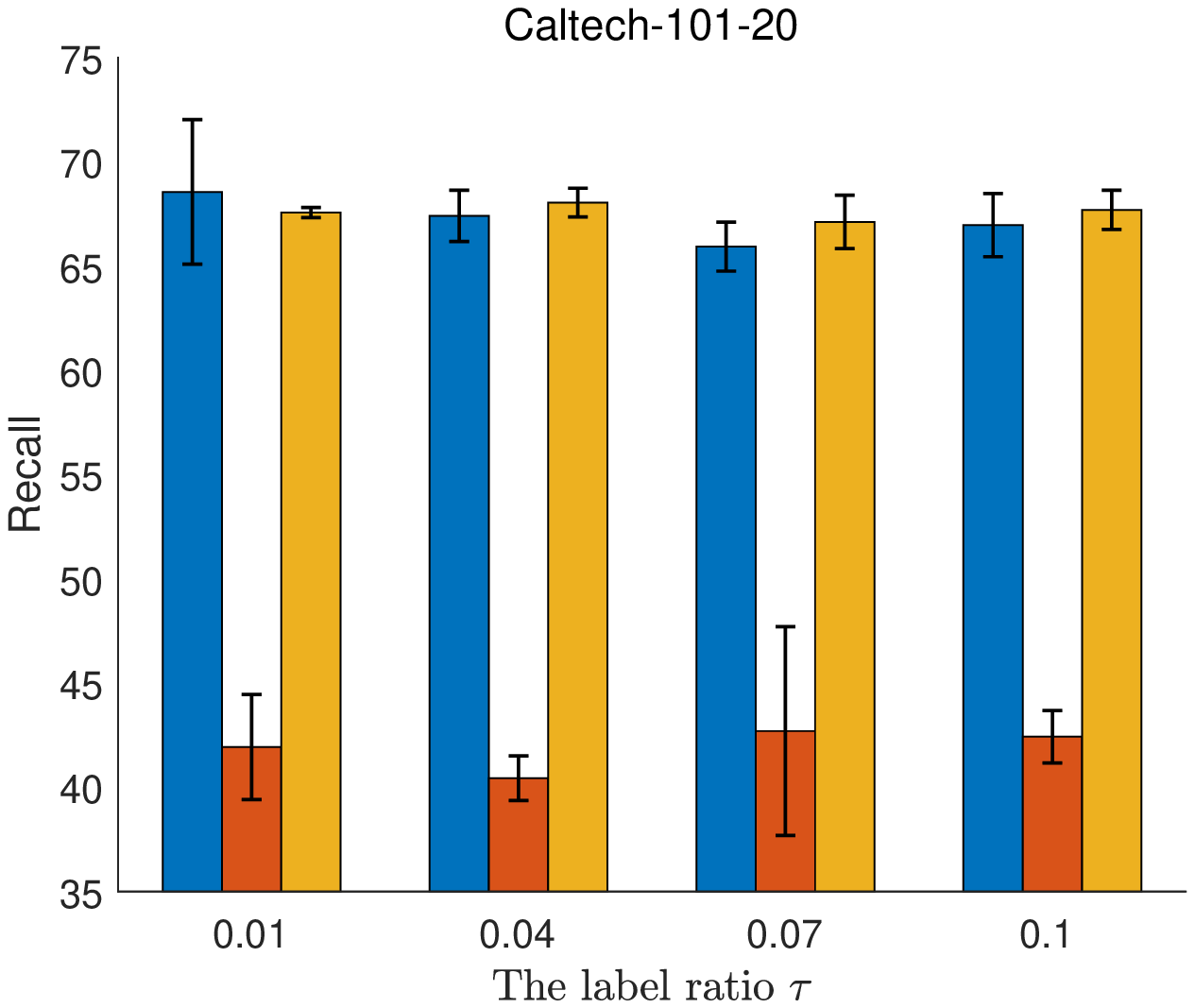}}
	\caption{The effect of the randomly selected training set in Caltech-101-20 dataset on the classification performance for the MLCK, DACK, and LACK methods.}
	\label{ranlabel_Cal-20_performance}
\end{figure*}

Figures \ref{ranlabel_Cal-7_performance} and \ref{ranlabel_Cal-20_performance} report the average classification performance and the corresponding standard deviation of the methods. The following points are known from the experimental results: The proposed LACK is able to achieve leading classification performance in most cases, which demonstrates the effectiveness of the label-driven auto-weighted strategy. The standard deviation of LACK in most cases is almost indistinguishable from that of MCCK with equal treatment of view weights, which demonstrates our proposed label-driven auto-weighted strategy that does not depend on the selection of training data.

\begin{figure*}[!t]
	\centering
	\subfigure[$\tau =0.01$.]{
		\includegraphics[width=0.23\textwidth]{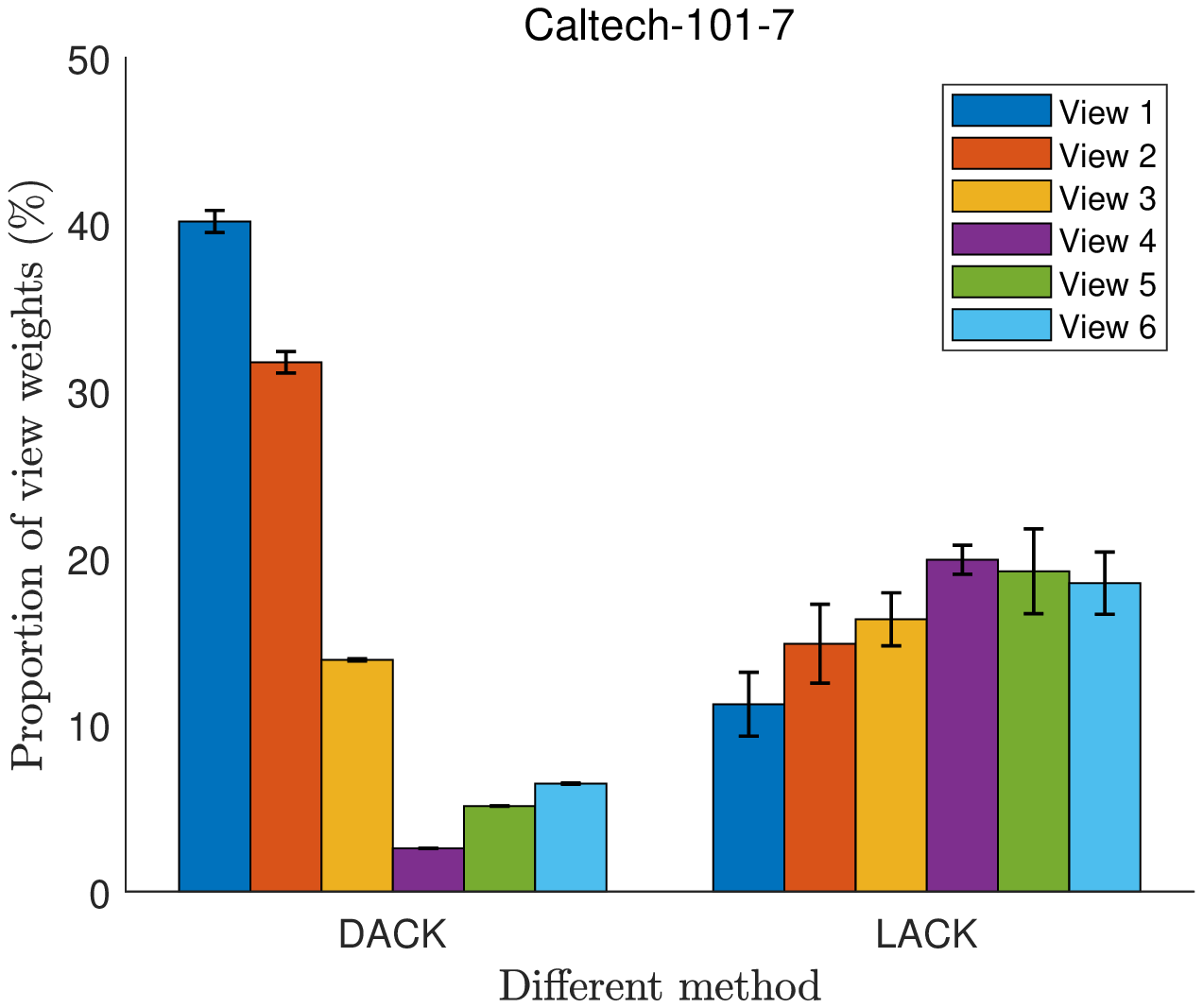}}
	\subfigure[$\tau =0.04$.]{
		\includegraphics[width=0.23\textwidth]{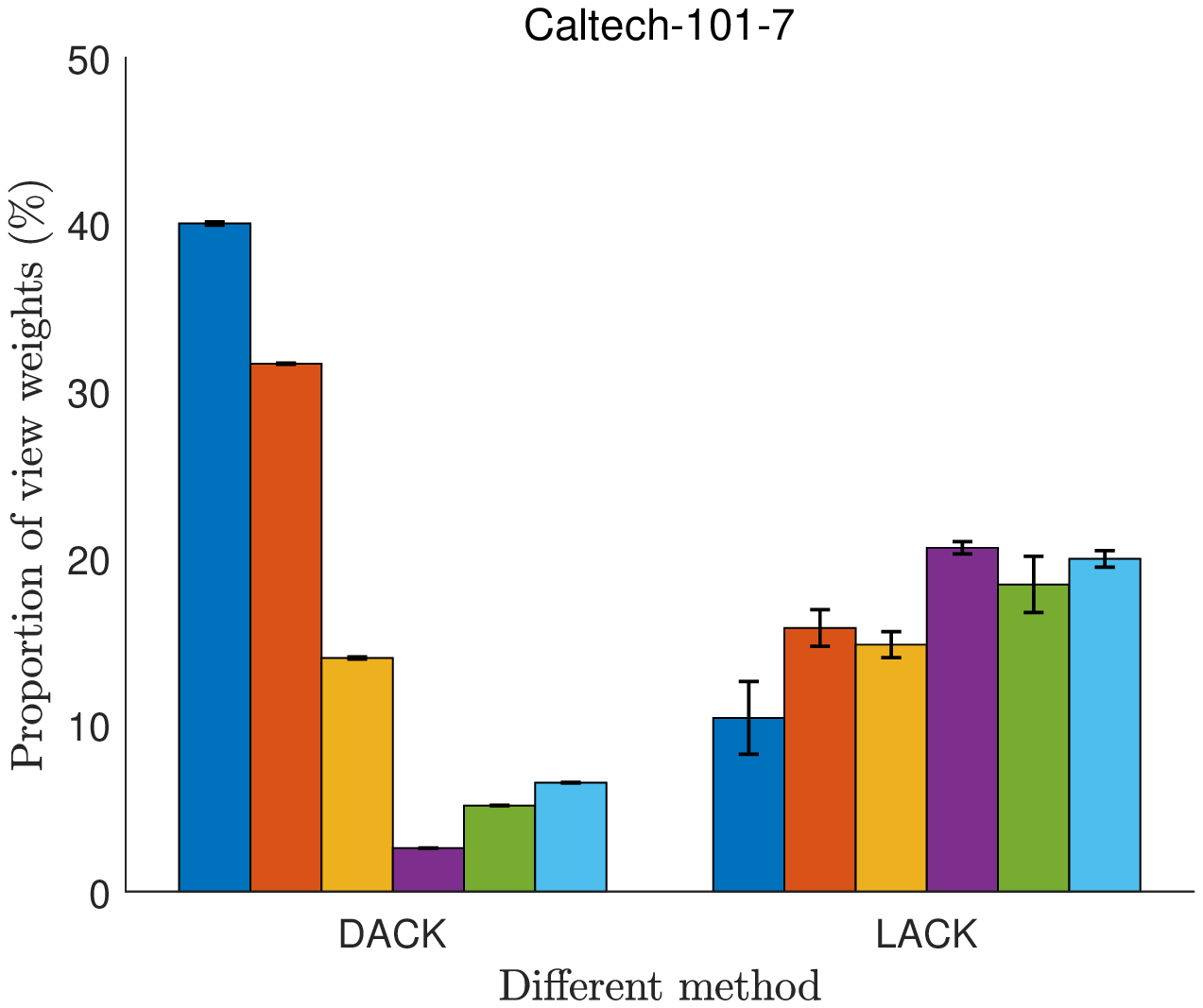}}
	\subfigure[$\tau =0.07$.]{
		\includegraphics[width=0.23\textwidth]{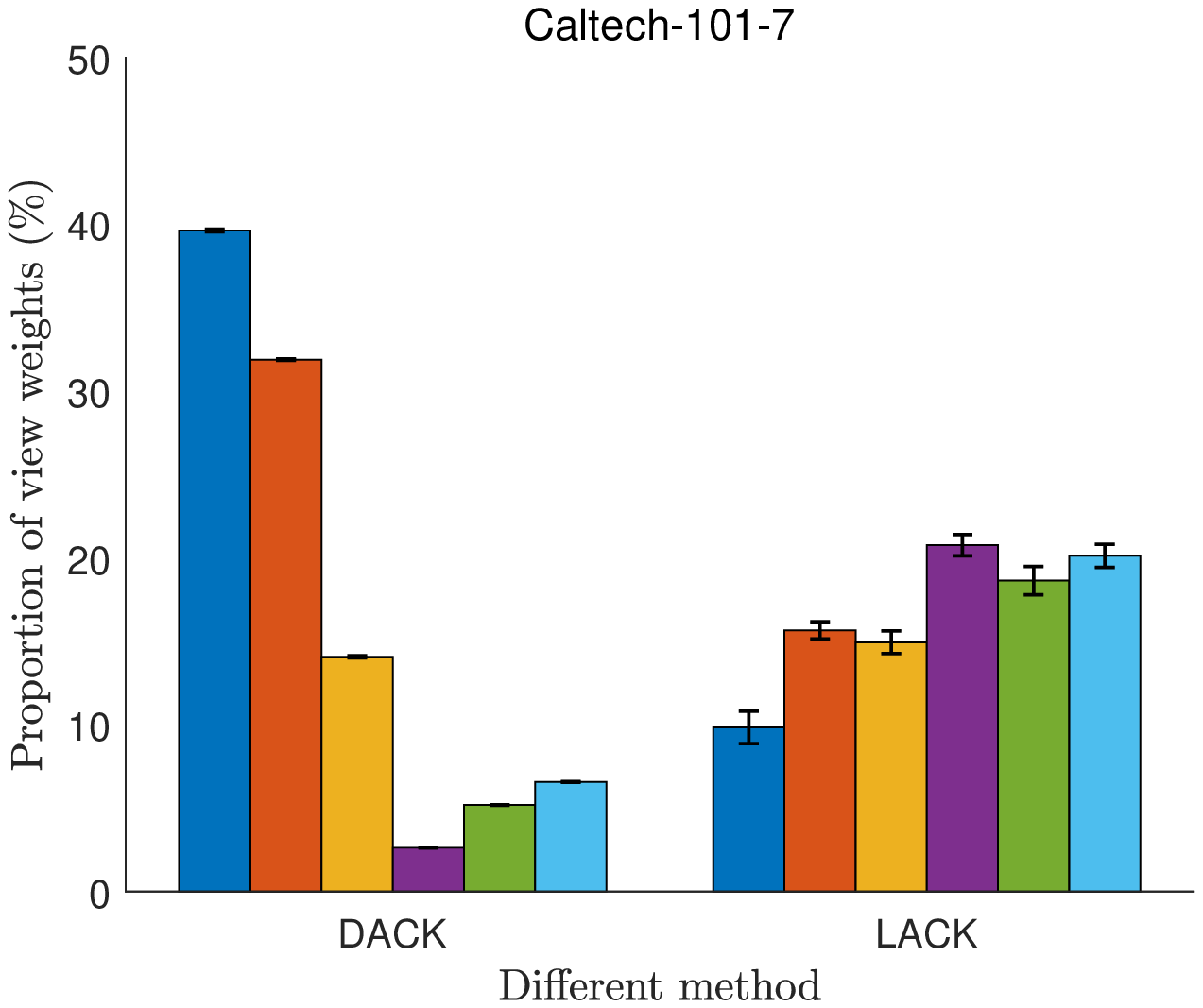}}
	\subfigure[$\tau =0.1$.]{
		\includegraphics[width=0.23\textwidth]{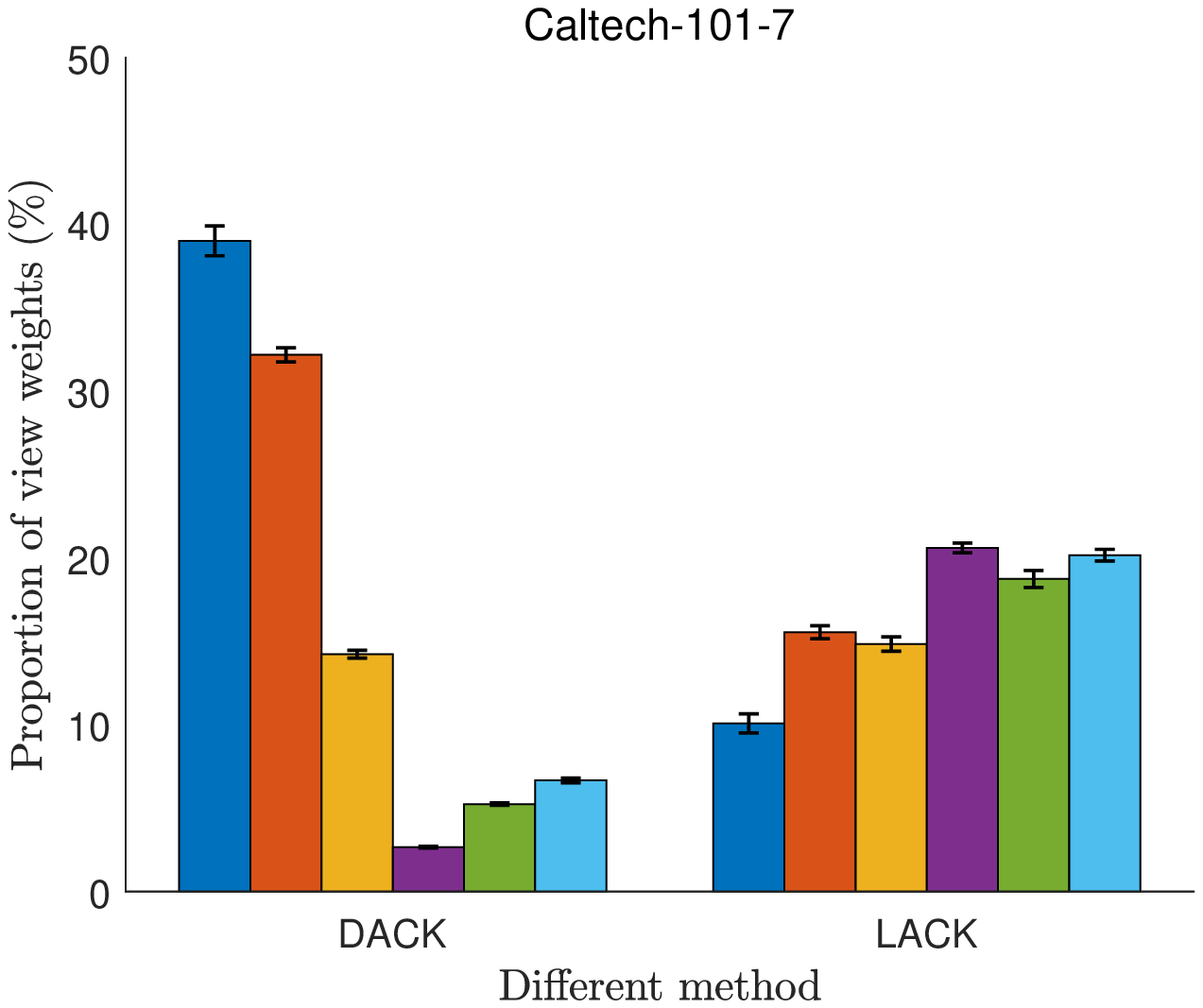}}
	\caption{The effect of the randomly selected training set in Caltech-101-7 dataset on the weight of views obtained by the DACK and LACK methods.}
	\label{ranlabel_Cal-7_weight}
\end{figure*}

\begin{figure*}[!t]
	\centering
	\subfigure[$\tau =0.01$.]{
		\includegraphics[width=0.23\textwidth]{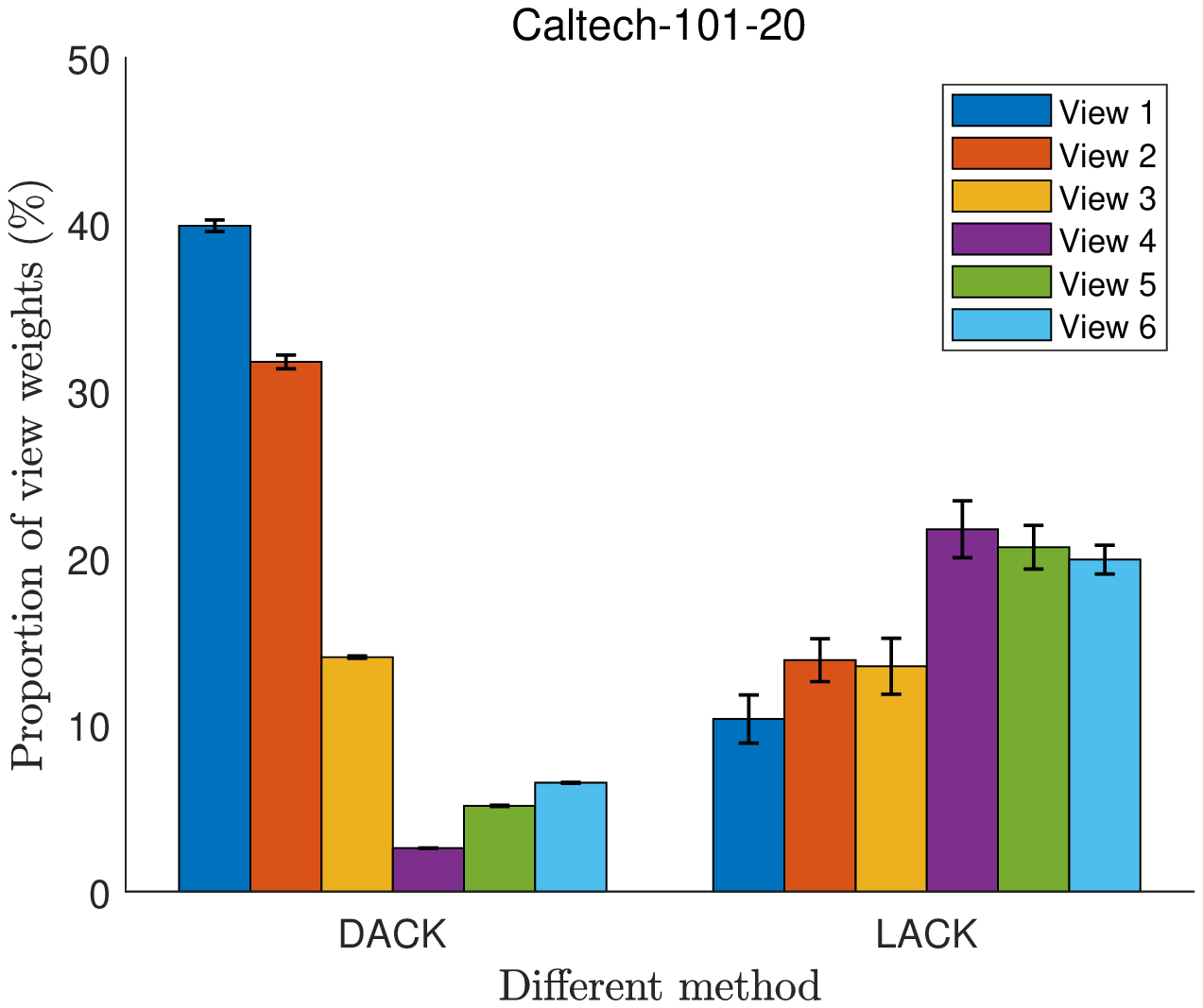}}
	\subfigure[$\tau =0.04$.]{
		\includegraphics[width=0.23\textwidth]{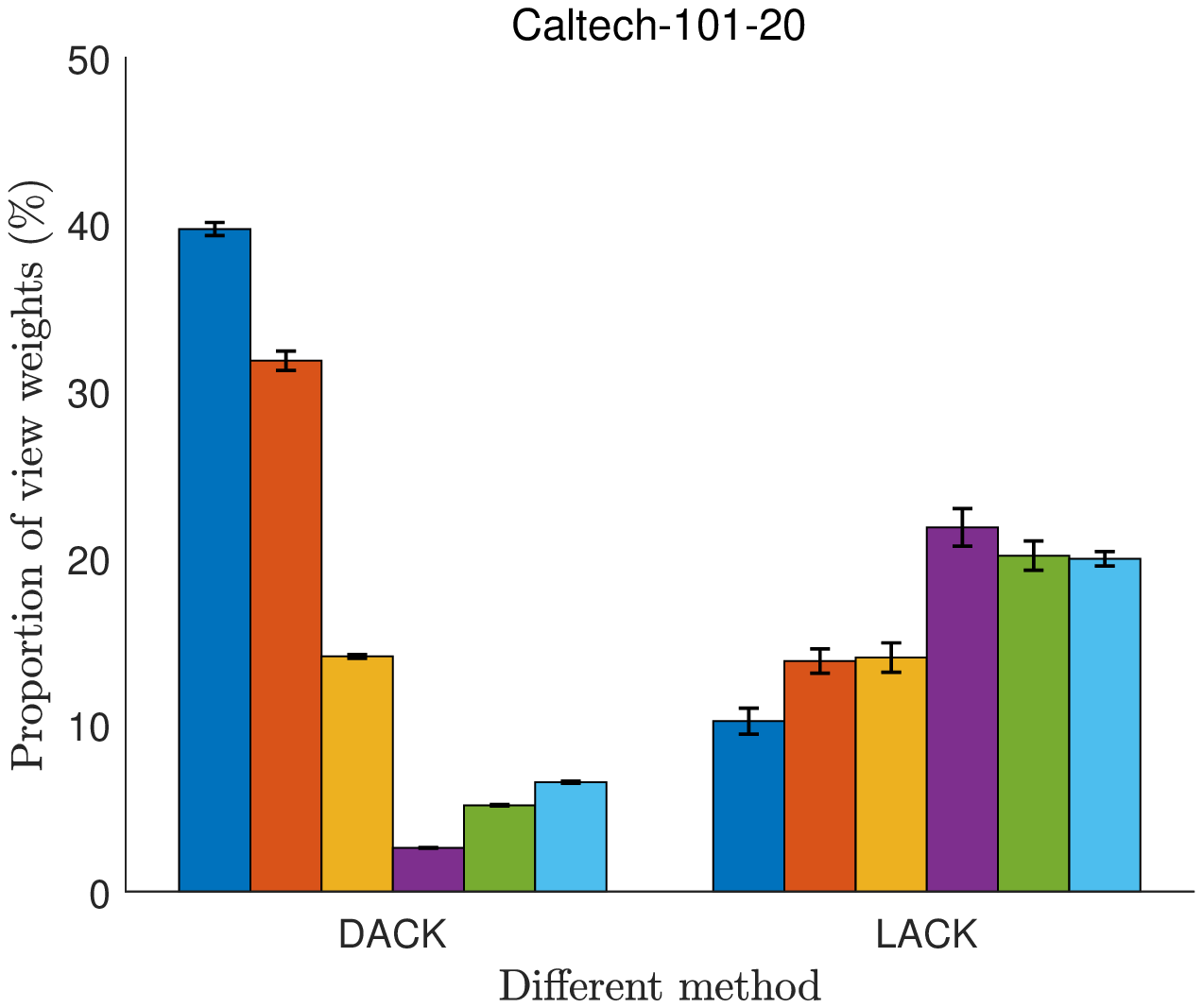}}
	\subfigure[$\tau =0.07$.]{
		\includegraphics[width=0.23\textwidth]{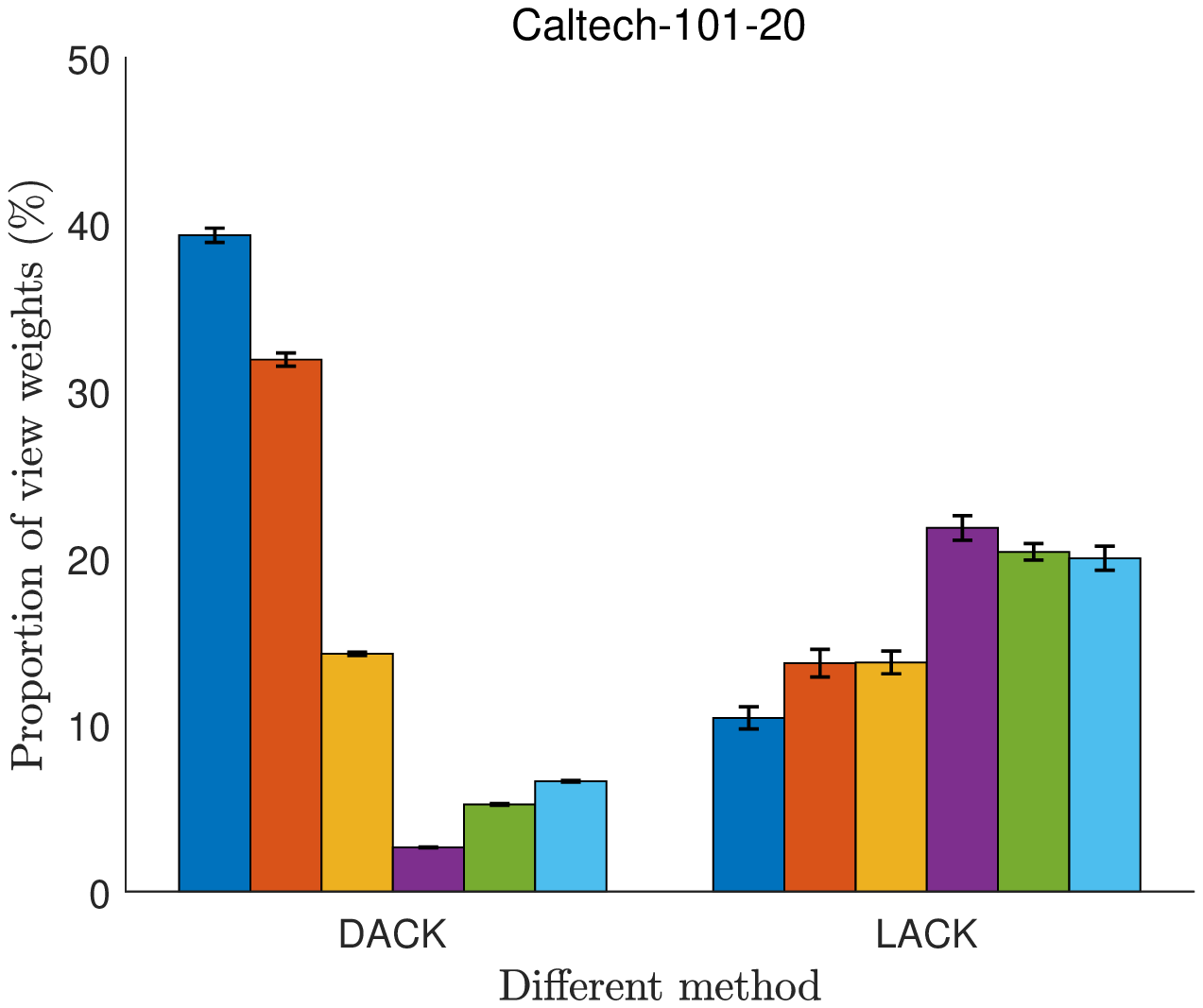}}
	\subfigure[$\tau =0.1$.]{
		\includegraphics[width=0.23\textwidth]{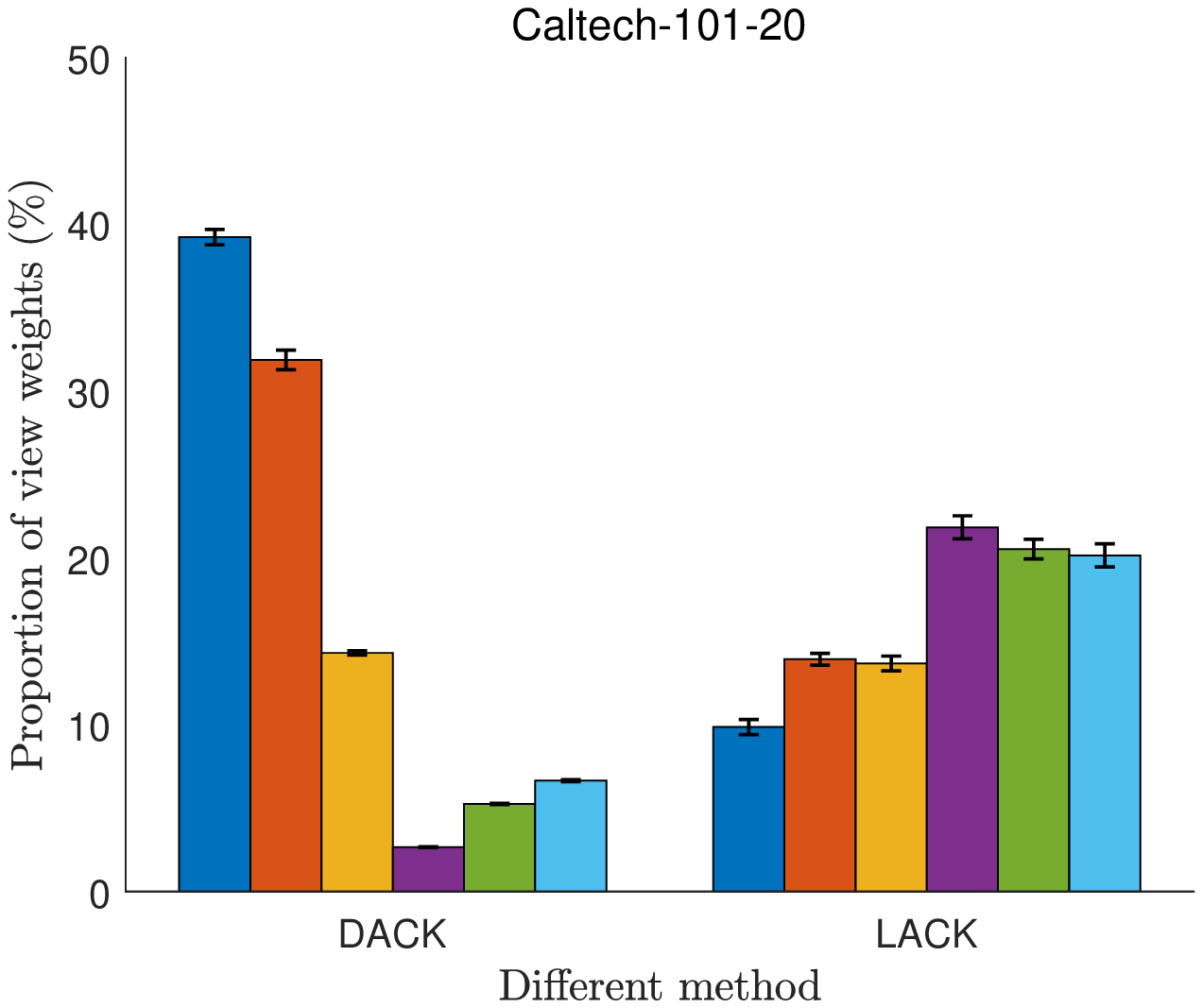}}
	\caption{The effect of the randomly selected training set in Caltech-101-20 dataset on the weight of views obtained by the DACK and LACK methods.}
	\label{ranlabel_Cal-20_weight}
\end{figure*}

Figures \ref{ranlabel_Cal-7_weight} and \ref{ranlabel_Cal-20_weight} report the mean and standard deviation of the view weights finally obtained by the method. From the results, it can be obtained that the view weights obtained by the proposed LACK relying on the training labels tend to be stable as the label ratio increases under different training data. Even if the labeling ratio is as low as $\tau =0.01$, the standard deviation of the view weights obtained by LACK is still satisfactory.

\subsection{Convergence Analysis}
\begin{figure}[!t]
	\centering  
	\subfigure[ ]{
		\includegraphics[width=0.23\textwidth]{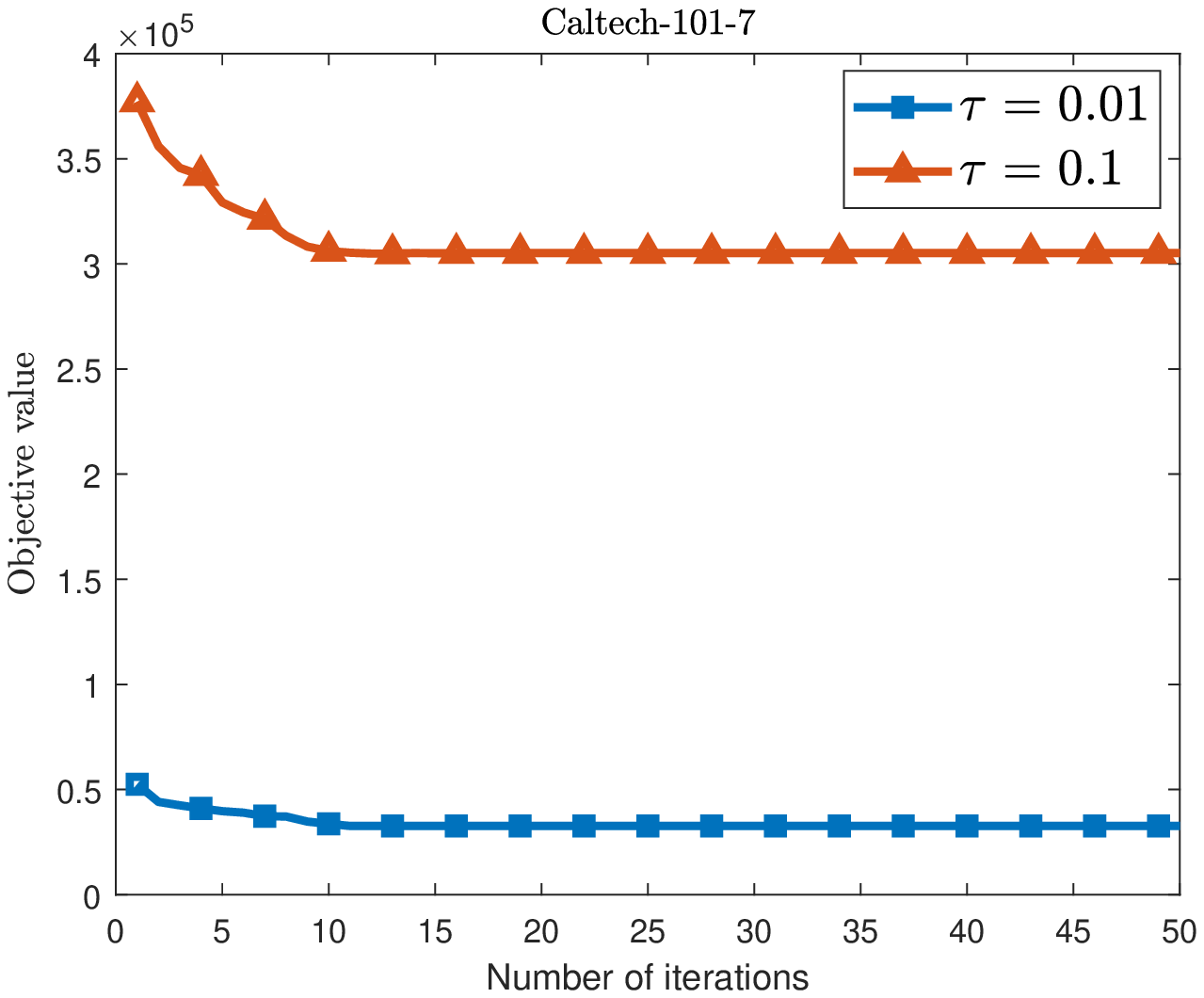}}\hspace{-2mm}
	\subfigure[ ]{
		\includegraphics[width=0.23\textwidth]{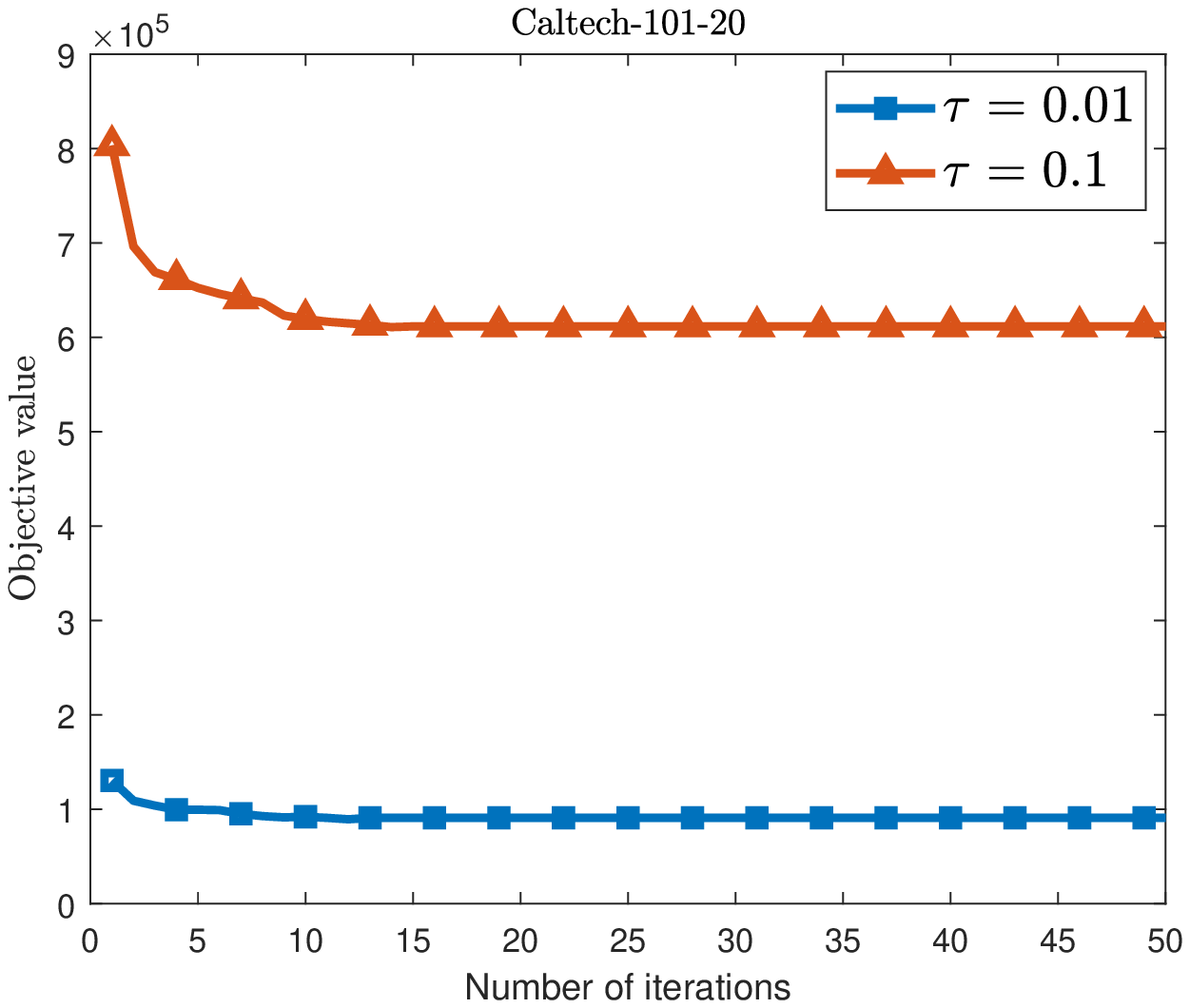}}
	\subfigure[ ]{
		\includegraphics[width=0.23\textwidth]{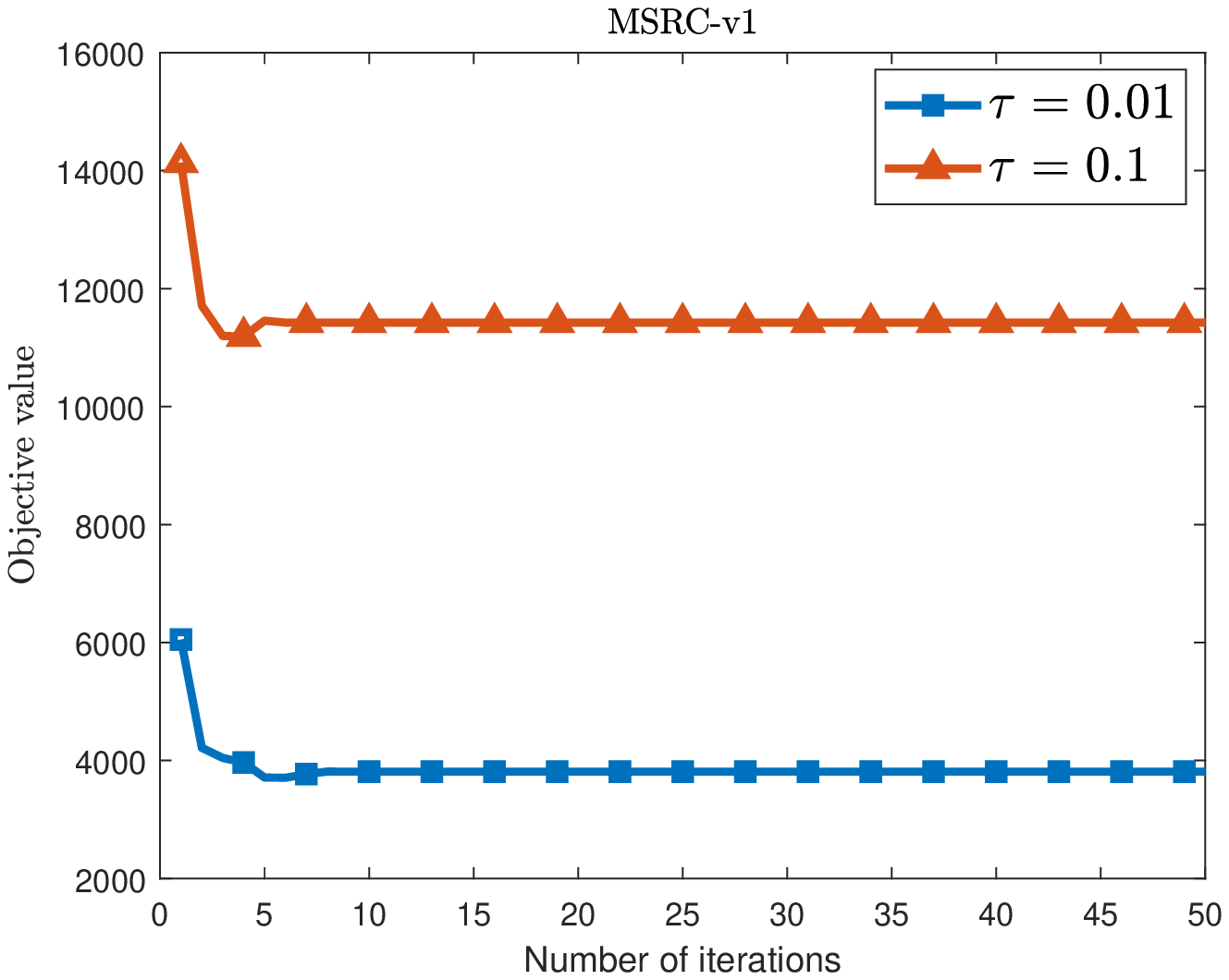}}\hspace{-2mm}
	\subfigure[ ]{
		\includegraphics[width=0.23\textwidth]{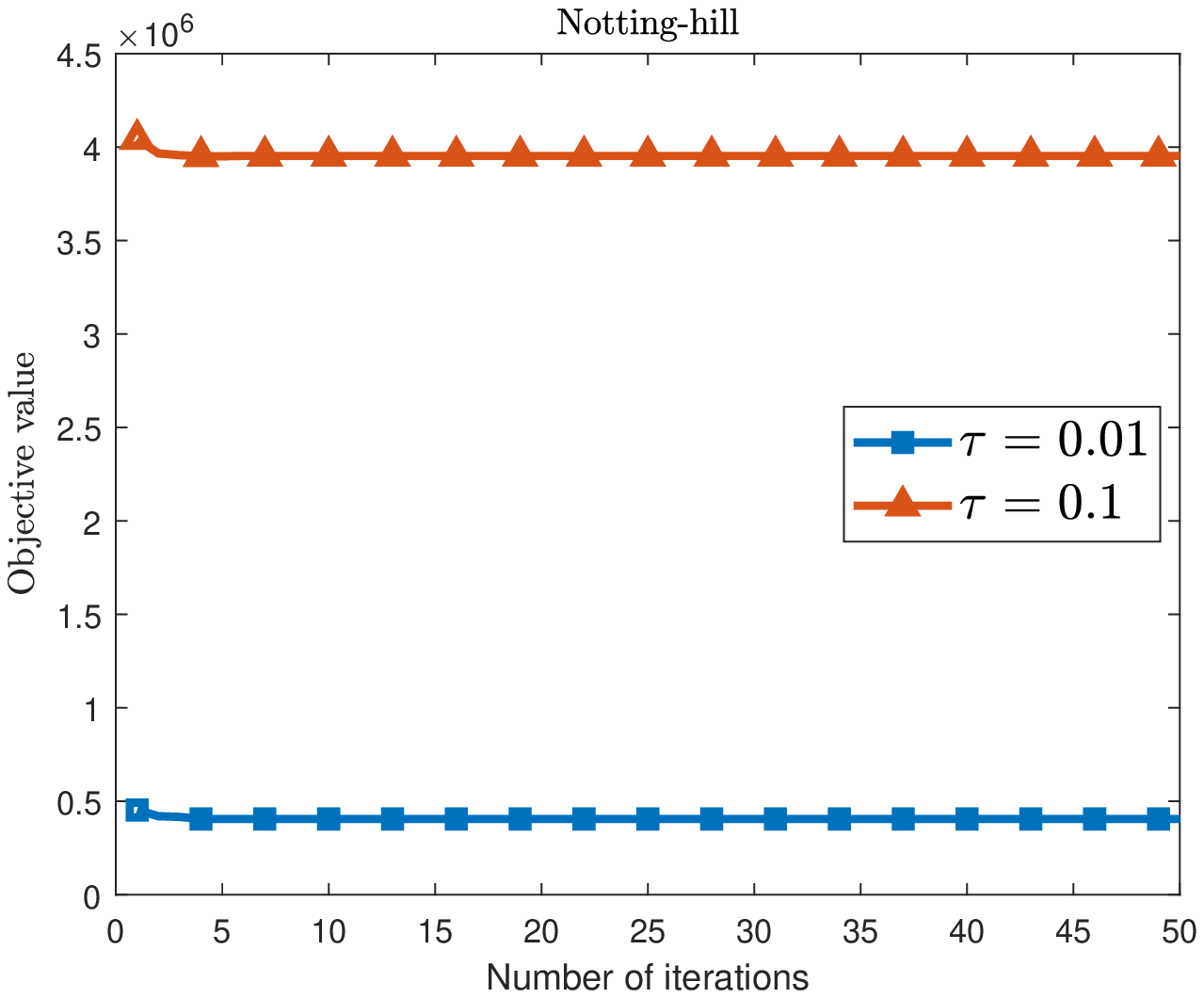}}
	\caption{Convergence curves of LACK in different label ratios $\tau=\{0.01,0.1\}$ on four datasets. (a) Caltech-101-7; (b) Caltech101-20; (c) MSRC-v1; (b) Noptting-hill.}
	\label{fig:F1}
\end{figure}

As discussed in Section \ref{convergence}, LACK can be theoretically guaranteed to converge to a local solution. To explore this point in more depth, we show objective values of LACK for 50 iterations in $\tau=\{0.01,0.1\}$ on the datasets, as shown in Table \ref{fig:F1}. From the experimental results, it can be seen that LACK is able to converge in all data sets in the different label ratios $\tau$.

\section{Conclusions}
\label{sec:5}
In this paper, we have investigated the problem of view importance distinguishing in multi-view data. We developed a label-driven auto-weighted strategy to distinguish view importance using label information as a priori. This strategy distinguishes view importance more accurately at a lower computational cost than traditional data-driven auto-weighted strategy. Moreover, this strategy is extremely scalable and can be easily extended to any kind of multi-view classification model with labels as hard constraints. Based on this strategy, we also proposed a transductive semi-supervised classification model to learn the labels of multi-view test data. The model can efficient optimization with guaranteed local convergence. The experimental results show that our model achieves encouraging performance both in terms of classification performance and computational efficiency, and can accurately distinguish the importance of different views.
\section*{Acknowledgments}
This work was supported in part by the National Natural Science Foundation of China under Grant 62073087 and Grant U1911401.

\bibliographystyle{IEEEtran}

\bibliography{MCACK_bib}

\end{document}